\documentclass[10pt,twocolumn,letterpaper]{article}

\usepackage[pagenumbers]{wacv}

\usepackage{graphicx}
\usepackage{amsmath}
\usepackage{amssymb}
\usepackage{booktabs}
\usepackage{colortbl}
\usepackage[table]{xcolor}
\usepackage{soul}

\usepackage[pagebackref,breaklinks,colorlinks]{hyperref}

\usepackage[capitalize]{cleveref}
\crefname{section}{Sec.}{Secs.}
\Crefname{section}{Section}{Sections}
\Crefname{table}{Table}{Tables}
\crefname{table}{Tab.}{Tabs.}

\def\wacvPaperID{2030}
\def\confName{WACV}
\def\confYear{2024}

\begin{document}

\title{MotionAGFormer: Enhancing 3D Human Pose Estimation with a Transformer-GCNFormer Network}

\author{Soroush Mehraban\textsuperscript{1,2}, Vida Adeli\textsuperscript{1,3}, Babak Taati\textsuperscript{1,2,3}\\
\textsuperscript{1}KITE Research Institute,
\textsuperscript{2}Institute of Biomedical Engineering, University of Toronto,\\
\textsuperscript{3}Department of Computer Science, University of Toronto\\
{\tt\small \{soroush.mehraban, vida.adeli\}@mail.utoronto.ca, babak.taati@uhn.ca}
}
\maketitle

\begin{abstract}
    Recent transformer-based approaches have demonstrated excellent performance in 3D human pose estimation. However, they have a holistic view and by encoding global relationships between all the joints, they do not capture the local dependencies precisely. In this paper, we present a novel Attention-GCNFormer (AGFormer) block that divides the number of channels by using two parallel transformer and GCNFormer streams. Our proposed GCNFormer module exploits the local relationship between adjacent joints, outputting a new representation that is complementary to the transformer output. By fusing these two representation in an adaptive way, AGFormer exhibits the ability to better learn the underlying 3D structure. By stacking multiple AGFormer blocks, we propose MotionAGFormer in four different variants, which can be chosen based on the speed-accuracy trade-off. We evaluate our model on two popular benchmark datasets: Human3.6M and MPI-INF-3DHP. MotionAGFormer-B achieves state-of-the-art results, with P1 errors of 38.4~mm and 16.2~mm, respectively. Remarkably, it uses a quarter of the parameters and is three times more computationally efficient than the previous leading model on Human3.6M dataset. Code and models are available at \url{https://github.com/TaatiTeam/MotionAGFormer}.
\end{abstract}

\begin{figure}[!t]
    \centering
    \includegraphics[width=0.45\textwidth]{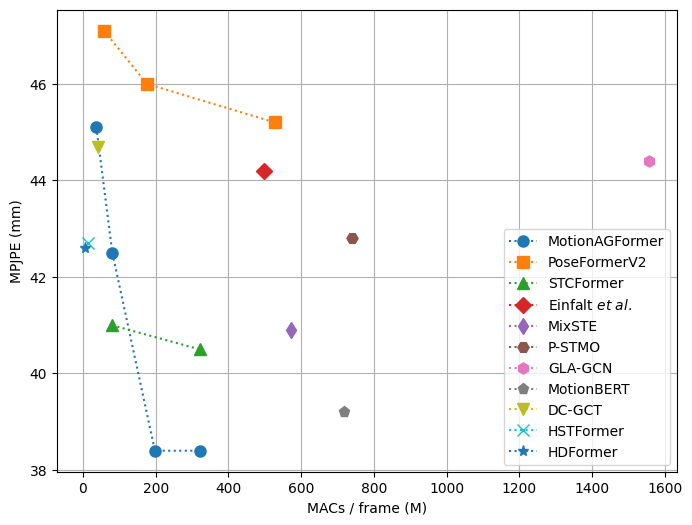}
    \caption{Comparisons of recent pose uplifting methods on Human3.6M~\cite{ionescu2013human3} (lower is better). MACs/frame denotes multiply-accumulate operations per each output frame. The proposed MotionAGFormer presents different variants and attains superior results, while maintaining computational efficiency.}
    \label{fig:macs-mpjpe}
\end{figure}

\section{Introduction}
\label{sec:intro}

Human pose estimation in 3D space is an active area of research with significant implications for numerous applications, from augmented~\cite{lin2010augmented} and virtual reality~\cite{mehta2017vnect} to autonomous vehicles~\cite{wiederer2020traffic, czech2022board, bauer2023weakly}, human-computer interaction~\cite{munea2020progress} and beyond. With this vast range of applications, the demand for more \textit{accurate} and \textit{computationally efficient} pose estimation models continues to grow. In most real-world scenarios, pose sequences are captured in 2D, primarily due to the prevalent use of standard RGB cameras. Consequently, one of the pivotal challenges in the field has been to effectively lift these 2D sequences into a 3D space. Accurate 3D human pose estimation enables the extraction of rich spatio-temporal information about human movements, and a deeper understanding of activities and interactions.
Recent 3D lifting models leverage the inherent spatial and temporal coherence of human movements to enhance the precision of 3D pose predictions. Nonetheless, despite the considerable advancements, there are several significant challenges that require attention.

The Transformer architecture~\cite{vaswani2017attention}, 
originally designed for NLP tasks, has been adapted to various computer vision problems, including pose estimation. Its ability to capture long-range dependencies and its innate self-attention mechanism make it a promising candidate for this domain. However, a sole reliance on global attention mechanisms, as employed by standard Transformers, may not be optimal for pose estimation tasks. Human motion is inherently structured with local spatial and temporal dependencies.

One primary concern is the modeling of skeleton relations over time. Existing methods predominantly rely either on transformer architectures or graph-based models. While transformers excel at capturing long-term dependencies, graph models excel at local dependencies. So, there is an opportunity for a unified architecture that integrates the global perspective of transformers with the local precision of graph models.

Additionally, the race for achieving SOTA accuracy has often led to the development of increasingly complex models with a large number of parameters. Such models, despite their good accuracy, often become impractical for real-world applications where computational efficiency and swift response times are pivotal.
Moreover, a predominant approach in recent models has been the prediction of a single 3D pose only for the central frame from a sequence of frames. This method, while seemingly efficient, leads to computational redundancy as it requires the reprocessing of several overlapping sequences. As a result, we instead employ a streamlined inference strategy that optimally exploits sequential data. This approach minimizes redundancy by predicting the complete 3D sequence of the input at a single forward pass.

In this paper, we introduce the MotionAGFormer, a novel transformer-graph hybrid architecture tailored for 3D human pose estimation. At its core, the MotionAGFormer harnesses the power of transformers to capture global information while simultaneously employing Graph Convolutional Networks (GCNs) to integrate local spatial and temporal relationships. We use an adaptive fusion to aggregate features from the transformer and graph streams. By doing so, we ensure a balanced and comprehensive representation of human motion, leading to enhanced accuracy in the 3D pose estimation (See Figure~\ref{fig:macs-mpjpe}).

In summary, the main contributions of our paper are: 
\begin{itemize}
  \setlength\itemsep{0em}
  \item \textit{Novel Design:} We propose the \textit{MotionAGFormer} model, in which we introduce a new GCNFormer module that excels in representing local dependencies inherent in human pose sequences.

  \item \textit{Efficiency and Flexibility:} \textbf{i)} Our MotionAGFormer stands out due to its lightweight nature and faster speed with fewer parameters compared to previous SOTA methods, without compromising on accuracy. \textbf{ii)} Recognizing diverse needs, we offer different variants of MotionAGFormer, granting users the flexibility to make a balanced choice between accuracy and speed based on their specific requirements.

  \item \textit{SOTA Performance:} MotionAGFormer achieves SOTA on two challenging datasets, Human3.6M and MPI-INF-3DHP.
\end{itemize}

\section{Related works}
\textbf{3D human pose estimation.}
Current approaches to tackle this problem can be understood from two perspectives. From one perspective, models can be categorized based on the input video, which can be either multi-view or monocular. Models that rely on multi-view inputs~\cite{zhang2021adafuse, reddy2021tessetrack, chun2023learnable, iskakov2019learnable, remelli2020lightweight} necessitate the simultaneous use of multiple cameras from different angles, which can be less feasible in real-world scenarios. From another perspective, considering their methodology, these models can be categorized as either direct 3D estimation approaches or 2D-3D lifting approaches. Direct 3D estimation methods~\cite{pavlakos2018ordinal, zhou2019hemlets, sun2018integral, pavlakos2017coarse} infer the joints in 3D coordinate from the video frames without any intermediate step. Inspired by the rapid development and availability of accurate 2D pose estimation models, more recently, 2D-3D lifting methods first use an off-the-shelf 2D pose detectors~\cite{cpn, hrnet, stackedhourglass} then lift 2D coordinates to 3D space~\cite{poseformer, poseformerv2, motionbert, STCFormer, mixste}. 
In this work, we are using 2D-3D lifting methods by having monocular video as the input.

\textbf{Transformer-based methods.} Transformers~\cite{vaswani2017attention} have shown promising result in different visual tasks~\cite{ViT, yu2022coca, su2023towards, zong2022detrs}. In the field of 3D human pose estimation, PoseFormer~\cite{poseformer} was the first purely transformer-based model. PoseFormerV2~\cite{poseformerv2} improved its computational efficacy by employing a frequency-domain representation that also made it robust against sudden movements in noisy data. MHFormer~\cite{li2022mhformer} addressed the problem of self-occlusion and depth ambiguity by proposing a module that learns multiple plausible pose hypotheses. P-STMO~\cite{pstmo} proposed masked pose modeling and reduced the final error by self-supervised pretraining the model. Enfalt \textit{et al.}~\cite{einfalt_up3dhpe_WACV23} decreased the computational complexity by leveraging masked token modeling. StridedFormer~\cite{li2022exploiting} replaced fully-connected layers in the feed-forward network of the transformer encoder with strided convolutions to progressively shrink the sequence length and efficiently lift the central frame. Unlike the abovementioned models that estimate the 3D pose  for only the center frame of a sequence, MixSTE~\cite{mixste} provided 3D estimates for every frame in the input sequence. STCFormer~\cite{STCFormer} decreased computational complexity by separating the correlation learning into spatial and temporal components. HSTFormer~\cite{qian2023hstformer} proposed hierarchical transformer encoders for better capturing spatial-temporal correlations. In addition to joint-joint attention which is commonly used, HDFormer~\cite{chen2023hdformer} included bone-joint and hyperbone-joint interactions. Some works exploited the output motion representation for various tasks. MotionBERT~\cite{motionbert} fine-tuned the model learned for the task of 3D human pose estimation for tasks such as action recognition and 3D human mesh recovery, while UPS~\cite{foo2023unified} trained a unified model for action recognition, 3D pose estimation, and early action prediction at the same time.

\textbf{Graph Convolutional Network.} GCN-based methods have achieved remarkable success within the domain of skeleton-based action recognition~\cite{msg3d, ctrn-gcn, hdgcn}. Despite their computational efficiency in 3D human pose estimation~\cite{choi2020pose2mesh, wang2020motion, zhao2019semantic, ci2019optimizing}, they usually cannot show competitive error compared to transformer-based counterparts. This is primarily due to their focus on local joints alone. Recently, GLA-GCN~\cite{yu2023glagcn} introduced an adaptive GCN approach to leverage global representation. By employing a strided design to reduce its temporal scope, they achieve competitive 3D human pose estimation against various transformer-based models, all while maintaining a lighter memory load.  
However, the effectiveness of the proposed module in extracting global representation is not on par with that of an attention module.

\textbf{Hybrid methods.} These methods use different modules to capture distinct aspects of the input sequence and are not extensively explored yet. Recently, DC-GCT~\cite{dc-gct} proposed a Local Constraint Module based on GCN, and Global Constraint Module based onf self-attention to exploit both local and global dependecies of the input sequence. However, as their model is designed to operate with both individual input frames and sequences of frames, it does not distinguish between temporal and spatial dimensions. As a result, it falls short in delivering competitive outcomes when contrasted with transformer-based methods.

\begin{figure*}
  \centering
  \includegraphics[width=13cm]{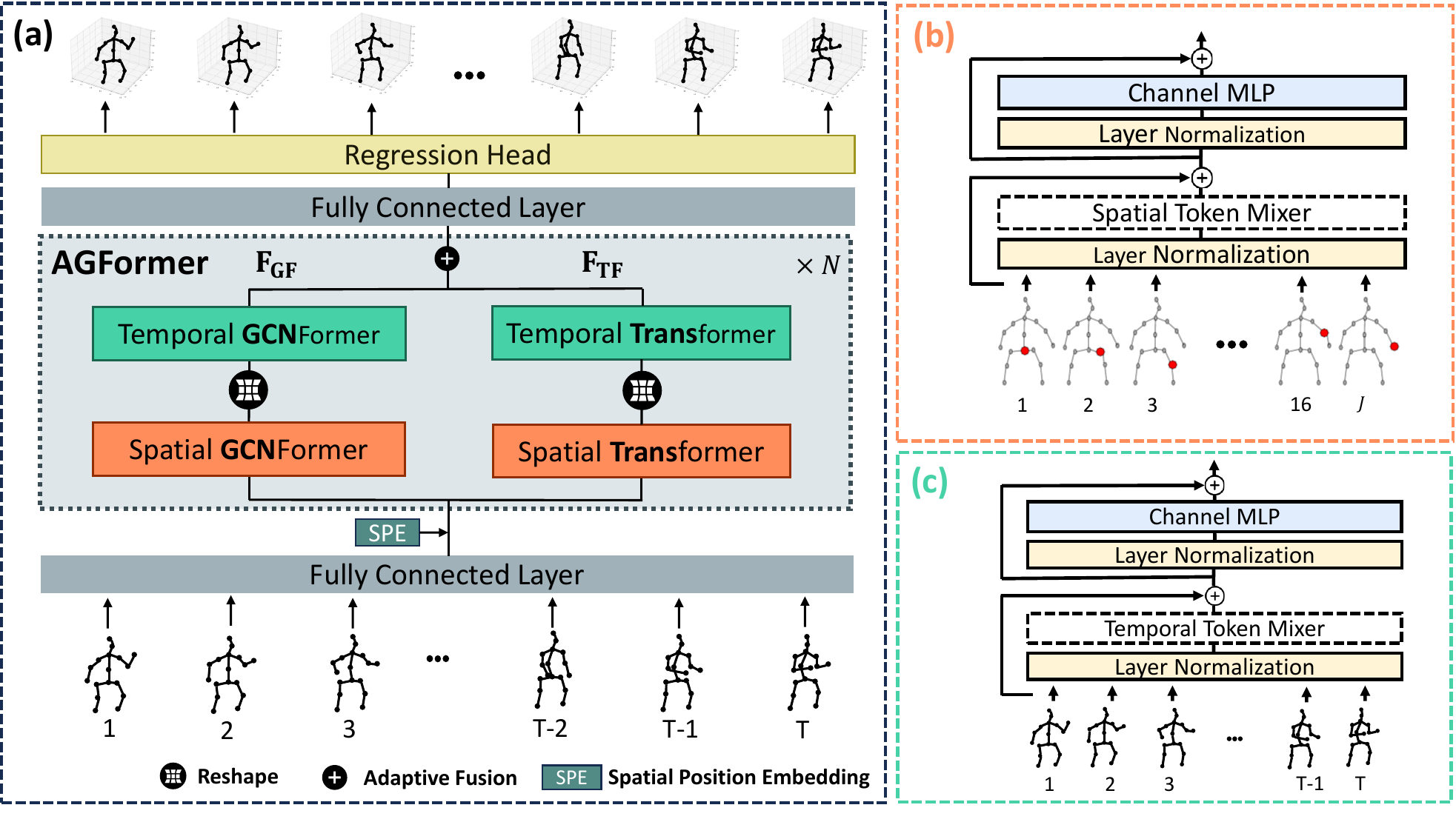}
  \caption{(a) \textbf{MotionAGFormer} is a novel architecture featuring $N$ dual-stream spatio-temporal blocks, wherein one stream employs Transformers and the other leverages GCNFormers. (b) \textbf{Spatial MetaFormer}. Each input token represents an individual joint of the human body. (c) \textbf{Temporal MetaFormer}. Input tokens are frames of pose sequence.}
  \label{fig:architecture}
\end{figure*}

\section{Method}
\subsection{Preliminary}
We begin this section by reviewing the concept of MetaFormer~\cite{yu2022metaformer}, which forms the core of our encoders. A MetaFormer can be described as a generalization of the Transformer architecture~\cite{vaswani2017attention}, wherein the attention module is substituted with any mixer capable of transforming information among tokens. Specifically, for an input $X \in \mathbb{R}^{N \times C}$, with $N$ denoting the token numbers and $C$ representing the embedding dimension, the token mixer can be formally expressed as
\begin{equation} \label{eq:tokenmixer}
    Y = \mathrm{TokenMixer}(\mathrm{Norm}(X)) + X,
\end{equation}
where $\mathrm{Norm}(\cdot)$ denotes a normalization method such as batch or layer normalization~\cite{ioffe2015batch, ba2016layer}, and $\mathrm{TokenMixer}(\cdot)$ denotes a module that combines information among tokens. Our approach uses two parallel token mixers: Multi-head Self-Attention (MHSA) and Graph Convolutional Networks (GCNs)~\cite{kipf2016gcn}, each contributing uniquely to the information transformation process.

\subsection{Overall architecture} \label{sec:overall-architecture}
    Our objective is to lift a 2D (pixel coordinate) skeleton sequence  to accurate 3D pose sequences. To this end, we propose the MotionAGFormer architecture, which uses both attention (Transformer) and graph convolutional (GCNFormer) to lift motion sequences. 
    An overview of this architecture is shown in Figure~\ref{fig:architecture}a. 
    
    The model takes a 2D input sequence with confidence score \mbox{$\mathbf{X} \in \mathbb{R}^{T \times J \times 3}$}, where $T$ and $J$ refer to the number of frames and joint numbers, respectively. It then proceeds to map each joint in each time frame to a $d$-dimensional feature, $\mathbf{F}^{(0)} \in \mathbb{R}^{T \times J \times d}$, using a linear projection layer. Then a spatial position embedding $\mathbf{P_{pos}^s} \in \mathbb{R}^{1 \times J \times d}$ is added to the tokens. It is important to highlight here that our model does not disregard the temporal token order as that information is preserved in the GCNFormer stream (further discussion in ablation studies~\ref{sec:ablation}). 
     
     Subsequent to position embedding, we use $N$ blocks of AGFormer (Section~\ref{sec:AGFormer}) to compute $\textbf{F}^{(i)} \in \mathbb{R}^{T \times J \times d}$ ($i = 1, ..., N$) to effectively capture the underlying 3D structure of the skeleton sequence. Finally, we map $\mathbf{F}^{(N)}$ to a higher dimension by applying a linear layer and $tanh$ activation to compute motion semantic $\textbf{M} \in \mathbb{R}^{T \times J \times d'}$ and use a regression head to estimate 3D pose $\mathbf{\hat{P}} \in \mathbb{R}^{T \times J \times 3}$. The lifting loss contains position ($L_{3D}$) and velocity ($L_{\Delta \mathbf{P}}$) terms defined as
    
    \begin{equation} \label{losses}
    \begin{aligned}
    L_{3D} &= \Sigma_{t=1}^{T}\Sigma_{j=1}^{J}\|\mathbf{\hat{P}}_{t, j} - \mathbf{P}_{t, j}\|, \\
    L_{\Delta \mathbf{P}} &= \Sigma_{t=2}^{T}\Sigma_{j=1}^{J}\|\Delta\mathbf{\hat{P}}_{t, j} - \Delta\mathbf{P}_{t, j}\|,
    \end{aligned}
    \end{equation}
    where $\Delta\mathbf{\hat{P}}_{t} = \mathbf{\hat{P}}_{t} - \mathbf{\hat{P}}_{t-1}$, $\Delta\mathbf{P}_{t} = \mathbf{P}_{t} - \mathbf{P}_{t-1}$. The total lifting loss is then defined as
    \begin{equation} \label{total-loss}
            L = L_{3D} + \lambda_{\Delta \mathbf{P}}L_{\Delta \mathbf{P}},
    \end{equation}
    where the constant coefficient $\lambda_{\Delta \mathbf{P}}$ is used to balance position accuracy and motion smoothness.

\subsection{AGFormer Block}
    \label{sec:AGFormer}
    The AGFormer block uses a dual-stream architecture. Each stream consists of two components: a Spatial MetaFormer (Figure~\ref{fig:architecture}b) followed by a Temporal MetaFormer (Figure~\ref{fig:architecture}c). The Spatial MetaFormer processes individual body joints as distinct tokens, effectively capturing intra-frame relationships within a single frame. The Temporal MetaFormer, on the other hand, treats each frame as a single token, thus capturing inter-frame relationships over time. The key distinction between the two streams lies in their token mixer type. While one stream employs Transformers, the other stream uses GCNFormers.
    
    \textbf{Transformer stream}. This stream employs a Spatial Multi-Head Self-Attention (S-MHSA) to capture spatial relationships, followed by a Temporal Multi-Head Self-Attention (T-MHSA) to capture temporal relationships. The S-MHSA is defined as
    \begin{equation} \label{eq:s-mhsa}
        \resizebox{.88\linewidth}{!}{$
        \begin{split}
            \text{S-MHSA}(\mathbf{Q_{s}}, \mathbf{K_{s}}, \mathbf{V_{s}}) = \mathrm{Concat(head_{i}, ..., head_{h})}\mathbf{W_{s}}^{(O)}, \\
            \mathrm{head_{i}} = \mathrm{softmax}(\frac{\mathbf{Q_{s}}^{(i)}(\mathbf{K_{s}}^{(i)})^T}{\sqrt{d_{k}}})\mathbf{V_{s}}^{(i)},
        \end{split}
        $}
    \end{equation}
    where $\mathbf{W_{s}}^{(O)}$ is a projection parameter matrix, $h$ is the number of parallel attention heads, and $d_{k}$ is the feature dimension of $\mathbf{K_{s}}$. For computing the query matrix $\mathbf{Q_{s}}$, the key matrix $\mathbf{K_{s}}$, and the value matrix $\mathbf{V_{s}}$, we have
    \begin{equation} \label{eq:s-mhsa-matrices}
        \resizebox{.88\linewidth}{!}{$
        \mathbf{Q_{s}}^i = \mathbf{F_{s}W_{s}}^{(Q, i)}, 
        \mathbf{K_{s}}^i = \mathbf{F_{s}W_{s}}^{(K, i)},
        \mathbf{V_{s}}^i = \mathbf{F_{s}W_{s}}^{(V, i)},
        $}
    \end{equation}
    where  $\mathbf{F_{s}} \in \mathbb{R}^{BT \times J \times d}$ is spatial feature and $\mathbf{W_{s}}^{(Q, i)}$, $\mathbf{W_{s}}^{(K, i)}$, $\mathbf{W_{s}}^{(V, i)}$ are projection matrices and $B$ is the batch size. The S-MHSA result is subsequently fed into a multilayer perceptron (MLP), followed by a residual connection and LayerNorm. This completes the first MetaFormer, i.e. the Spatial Transformer. 
    
    Next, we reshape $\mathbf{F_{s}}$ into $\mathbf{F_{T}} \in \mathbb{R}^{BJ \times T \times d}$ to prepare per-joint temporal feature as the input of T-MHSA. Here we have
    \begin{equation} \label{eq:t-mhsa}
        \resizebox{.88\linewidth}{!}{$
        \begin{split}
            \text{T-MHSA}(\mathbf{Q_{T}}, \mathbf{K_{T}}, \mathbf{V_{T}}) = \mathrm{Concat(head_{i}, ..., head_{h})}\mathbf{W_{T}}^{(O)}, \\
            \mathrm{head_{i}} = \mathrm{softmax}(\frac{\mathbf{Q_{T}}^{(i)}(\mathbf{K_{T}}^{(i)})^T}{\sqrt{d_{k}}})\mathbf{V_{T}}^{(i)},
        \end{split}
        $}
    \end{equation}
    where $\mathbf{Q_{T}}$, $\mathbf{K_{T}}$, and $\mathbf{V_{T}}$ are calculated similar to Eqn.~(\ref{eq:s-mhsa-matrices}).
    
    \textbf{GCNFormer stream}. Unlike the Transformer stream, which aggregates global information, the GCNFormer stream focuses on local spatial and temporal relationships present within the skeleton sequence. While the local information is also available to the Transformers, the inclusion of this parallel stream allows the model to more effectively balance the integration of local and global information (see ablation analysis~\ref{sec:ablation}). The customized GCN module~\cite{luo2022learning} used in our GCNFormer is defined as:
    \begin{equation} \label{eq:gcnformer-1}
        \resizebox{.88\linewidth}{!}{$
        GCN(\mathbf{F^{(i)}}) = \sigma(V^{l} + \mathrm{Norm}(\Tilde{D}^{-\frac{1}{2}}\Tilde{A}\Tilde{D}^{-\frac{1}{2}}\mathbf{F^{(i)}}W_{1} + \mathbf{F^{(i)}}W_{2}).
        $}
    \end{equation}
    Where $\Tilde{A}=A + I_{N}$ represents the adjacency matrix with self-connections added, $I_{N}$ stands for the identity matrix, $\Tilde{D}{ii} = \Sigma{j}\Tilde{A}{jj}$ is defined as the sum of elements along the diagonal of $\Tilde{A}$, and $W_{1}$, $W_{2}$ denote trainable weight matrices specific to each layer. The activation function $\sigma(\cdot)$, such as ReLU, is applied, along with Batch Normalization~\cite{ioffe2015batch}. The GCN's output is then passed through an MLP, followed by residual connection and LayerNorm.
    
\begin{figure}[t]
      \includegraphics[width=0.5\textwidth]{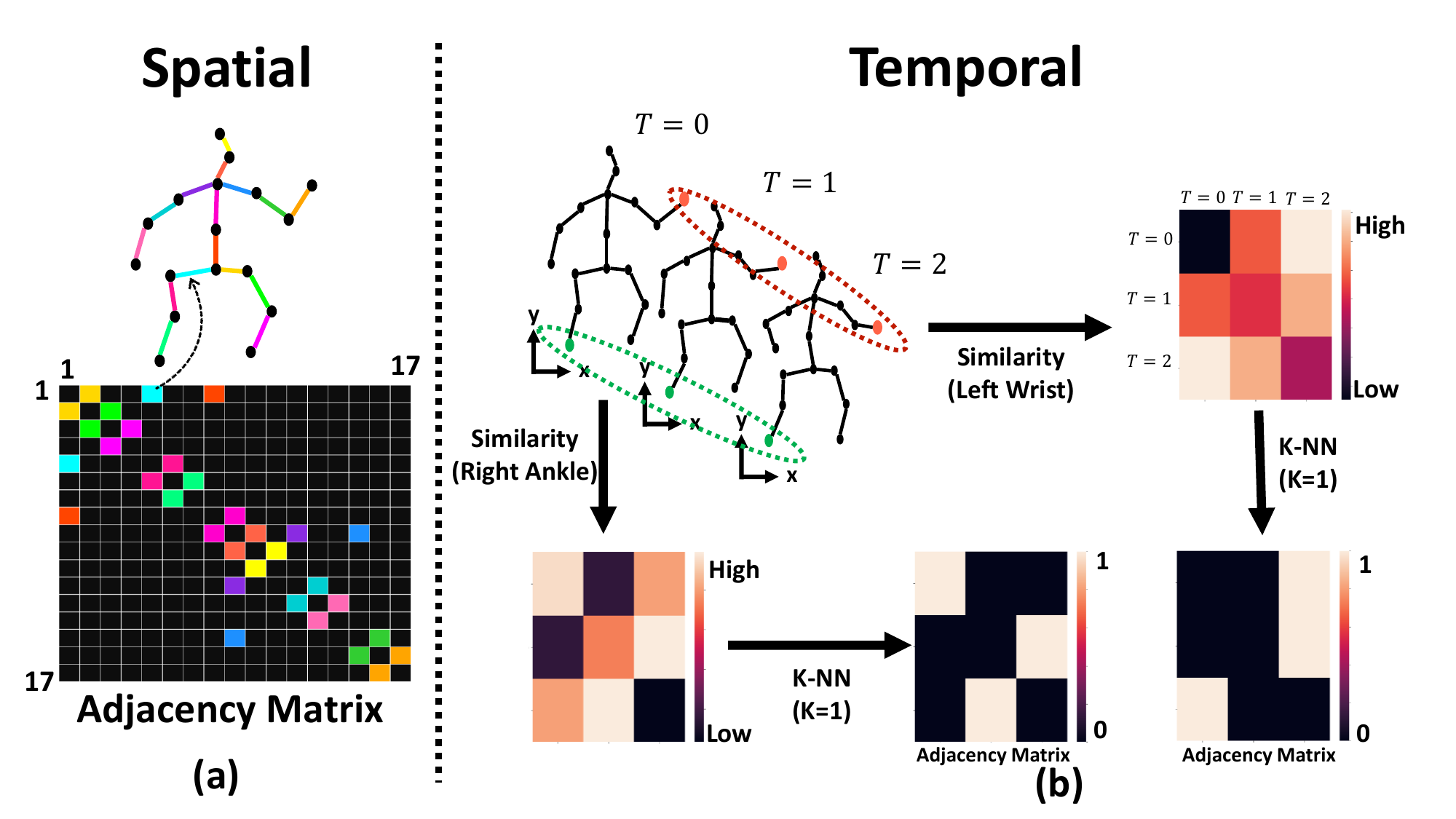}
      \caption{GCNFormer module topology. (a) The Spatial GCNFormer employs the human skeleton as its underlying topology. (b) The Temporal GCNFormer uses K-nearest neighbor (K-NN) to determine connected edges by considering the highest similarity of each joint (e.g., left wrist and right ankle in the figure) across the entire time frame. After K-NN, each row is connected to K columns.}
      \label{fig:graphtopology}
    \end{figure}

    The difference between the Spatial GCNFormer and the Temporal GCNFormer lies in their adjacency matrices and input features. The input features resemble that of the Transformer stream. In the Spatial GCNFormer, the adjacency matrix represents the human topology (Figure~\ref{fig:graphtopology}a). For Temporal GCNFormer, on the other hand, we calculate the similarity between a single joint at different time frames using 
    $Sim(\mathbf{F_{T}}^{t_{i}}, \mathbf{F_{T}}^{t_{j}}) = (\mathbf{F_{T}}^{t_{i}})^T\mathbf{F_{T}}^{t_{j}}$ and choose the $K$ nearest neighbors as the connected nodes in the graph (Figure~\ref{fig:graphtopology}b). Hence, the graph topology in Temporal GCNFormer is determined by the learned node features.
    
    \textbf{Adaptive Fusion}. Similar to MotionBERT~\cite{motionbert}, we use adaptive fusion to aggregate extracted features of the Transformer and GCNFormer streams. This is defined as:
    \begin{equation} \label{eq:adaptivefusion}
        \resizebox{.88\linewidth}{!}{$
        \mathbf{F}^{(i)} = {\alpha_{TF}}^{(i)}\circ \mathbf{F_{TF}}^{(i-1)} + {\alpha_{GF}}^{(i)}\circ \mathbf{F_{GF}}^{(i-1)}
        $},
    \end{equation}
    where $\mathbf{F}^{(i)}$ represents the feature embedding extracted at depth $i$, the element-wise multiplication denoted by $\circ$, and $\mathbf{F_{TF}}^{(i-1)}$, $\mathbf{F_{GF}}^{(i-1)}$ refer to the extracted Transformer stream and GCNFormer stream features at depth $i-1$, respectively. The adaptive fusion weights $\alpha_{TF}$ and $\alpha_{GF}$ are defined as
    \begin{equation} \label{eq:adaptivefusionweights}
        \resizebox{.88\linewidth}{!}{$
         {\alpha_{TF}}^{(i)}, {\alpha_{GF}}^{(i)} = \mathrm{softmax}(W \cdot \mathrm{Concat}(\mathbf{F_{TF}}^{(i-1)}, \mathbf{F_{GF}}^{(i-1)}))
        $},
    \end{equation}
    where $W$ is a learnable linear transformation.

\section{Experiments}
    We evaluate the performance of our proposed MotionAGFormer on two large-scale 3D human pose estimation datasets, i.e., Human3.6M~\cite{ionescu2013human3} and MPI-INF-3DHP~\cite{mehta2017monocular}.
    \subsection{Datasets and Evaluation Metrics}
    \textbf{Human3.6M} is a widely used indoor dataset for 3D human pose estimation. It contains 3.6 million video frames of 11 subjects performing 15 different daily activities. To ensure fair evaluation, we follow the standard approach and train the model using data from subjects 1, 5, 6, 7, and 8, and then test it on data from subjects 9 and 11. Following previous works~\cite{poseformerv2, STCFormer}, we use two protocols for evaluation. The first protocol (referred to as P1) uses Mean Per Joint Position Error (MPJPE) in millimeters between the estimated pose and the actual pose, after aligning their root joints (sacrum). The second protocol (referred to as P2) measures Procrustes-MPJPE, where the actual pose and the estimated pose are aligned through a rigid transformation.
    
\begin{figure*}
  \centering
  \includegraphics[width=15cm]{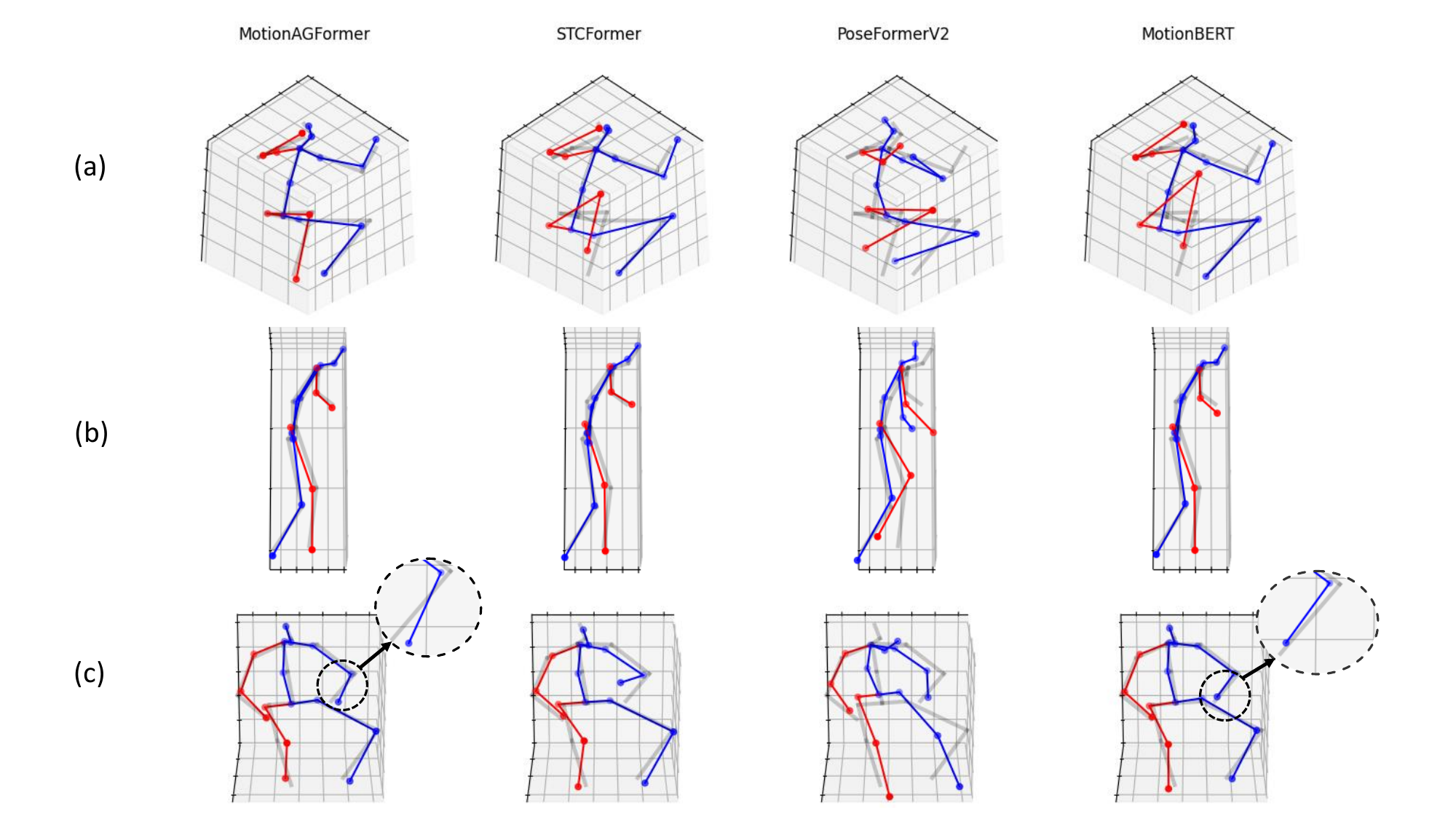}
  \caption{Qualitative comparisons of MotionAGFormer with STCformer~\cite{STCFormer}, PoseFormerV2~\cite{poseformerv2}, and MotionBERT~\cite{motionbert}. The transparent gray skeleton is the ground-truth 3D pose. Blue represents the torso and left part, and Red indicates the right part of the estimated body.}
  \label{fig:qualitative-comparison}
\end{figure*}

    \textbf{MPI-INF-3DHP} is another large-scale dataset gathered in three different settings: green screen, non-green screen, and outdoor environments. Following  previous works~\cite{poseformerv2, STCFormer}, MPJPE, Percentage of Correct Keypoint (PC) within 150\,mm range, and Area Under the Curve (AUC) are reported as evaluation metrics.
    \subsection{Implementation Details}
    \textbf{Model Variants.} We build four different configurations of our model, as summarized in Table~\ref{tab:variants}. Our base model, known as  MotionAGFormer-B, strikes a balance between accurate estimation and computational cost. The remaining variants are named according to their parameter size and computational demands and selection of each variant can be based on an application's requirements, such as choosing between real-time processing or more precise estimations. The motion semantic dimension is $d' = 512$, the expansion layer of each MLP is $\alpha = 4$, the number of attention heads is $h = 8$, and the number of temporal neighbours in GCNformer stream is $ k = 2$, for all experiments.

    \begin{table}[h]\small
        \caption{Details of MotionAGFormer model variants. $N$: Number of layers. $d$: Hidden size. $T$: Number of input frames.}
            \begin{tabular}{lccccc}
                \toprule
                Method & $N$ & $d$ & $T$ & Params & MACs \\
                \midrule
                MotionAGFormer-XS & 12 & 64 & 27 & 2.2 M & 1.0 G \\
                MotionAGFormer-S & 26 & 64 & 81 & 4.8 M & 6.6 G \\
                MotionAGFormer-B & 16 & 128 & 243 & 11.7 M & 48.3 G\\
                MotionAGFormer-L & 26 & 128 & 243 & 19.0 M & 78.3 G \\
                \bottomrule
            \end{tabular}%
        \label{tab:variants}
    \end{table}

    \textbf{Experimental settings.} Our model is implemented using PyTorch~\cite{pytorch} and executed on a setup with two NVIDIA A40 GPUs. We apply horizontal flipping augmentation for both training and testing following~\cite{motionbert, poseformerv2}. For model training, we set each mini-batch as 16 sequences. The network parameters are optimized using AdamW~\cite{AdamW} optimizer over 90 epochs with a weight decay of 0.01. The initial learning rate is set to 5e-4 with an exponential learning rate decay schedule and the decay factor is 0.99. We use the Stacked Hourglass~\cite{newell2016stacked} 2D pose detection results and 2D ground truths on Human3.6M, following~\cite{motionbert}. For MPI-INF-3DHP, ground truth 2D detection is used following a similar approach used in comparison baselines~\cite{poseformerv2, STCFormer}.
    
    \begin{table*}[h]\small
    \caption{Quantitative comparisons on Human3.6M.
    $T$: Number of input frames. CE: Estimating center frame only. P1: MPJPE error (mm). P2: P-MPJPE error (mm). ${\mathrm{P1}^\dag}$: P1 error on 2D ground truth. (*) denotes using HRNet~\cite{hrnet} for 2D pose estimation. The best and second-best scores are in bold and underlined, respectively. For per action result, refer to the supplementary material. }
      \centering
        \begin{tabular}{lcc|ccccc}
          \hline
          Method & $T$ & CE & Param & MACs & MACs/frame & P1$\downarrow$	
    /P2$\downarrow$	 & ${\mathrm{P1}^\dag}$$\downarrow$	 
     \\
          \hline
          *MHFormer~\cite{li2022mhformer} CVPR'22 & 351 & \checkmark & 30.9 M & 7.0 G & 7,096 M & 43.0/34.4 & 30.5 \\
          MixSTE~\cite{mixste} CVPR'22 & 243 & $\times$ & 33.6 M & 139.0 G & 572 M & 40.9/32.6 & 21.6\\
          P-STMO~\cite{pstmo} ECCV'22 & 243 & \checkmark & 6.2 M & 0.7 G & 740 M & 42.8/34.4 & 29.3\\
          Stridedformer~\cite{li2022exploiting} TMM'22 & 351 & \checkmark & 4.0 M & 0.8 G & 801 M &43.7/35.2 & 28.5\\
          Einfalt~\textit{et al.}~\cite{einfalt_up3dhpe_WACV23} WACV'23 & 351 & \checkmark & 10.4 M & 0.5 G & 498 M & 44.2/35.7 & - \\
          STCFormer~\cite{STCFormer} CVPR'23  & 243 & $\times$ & 4.7 M & 19.6 G & 80 M & 41.0/\underline{32.0} & 21.3 \\
          STCFormer-L~\cite{STCFormer} CVPR'23 & 243 & $\times$ & 18.9 M & 78.2 G & 321 M & 40.5/\textbf{31.8} & - \\
          PoseFormerV2~\cite{poseformerv2} CVPR'23 & 243 & \checkmark & 14.3 M & 0.5 G & 528 M & 45.2/35.6 & -\\
          UPS~\cite{foo2023unified} CVPR'23 & 243 & \checkmark & - & - & - & 40.8/32.5 & -\\
          GLA-GCN~\cite{yu2023glagcn} ICCV'23 & 243 & \checkmark & 1.3 M & 1.5 G & 1,556 M & 44.4/34.8 & 21.0\\
          MotionBERT~\cite{motionbert} ICCV'23 & 243 & $\times$ & 42.5 M & 174.7 G & 719 M & \underline{39.2}/32.9 & \underline{17.8}\\
          HDFormer~\cite{chen2023hdformer} IJCAI'23 & 96 & $\times$ & 3.7 M & 0.6 G & 6 M & 42.6/33.1 & 21.6\\
          HSTFormer~\cite{qian2023hstformer} arXiv'23 & 81 & $\times$ & 22.7 M & 1.0 G & 13 M & 42.7/33.7 & 27.8\\
          DC-GCT~\cite{dc-gct} arXiv'23 & 81 & \checkmark & 3.1 M & 41 M & 41 M & 44.7/- & -\\
          \rowcolor{gray!10} MotionAGFormer-XS & 27 & $\times$ & 2.2 M & 1.0 G & 37 M & 45.1/36.9 & 28.1 \\
          \rowcolor{gray!10} MotionAGFormer-S & 81 & $\times$ & 4.8 M & 6.6 G & 81 M & 42.5/35.3 & 26.5 \\
          \rowcolor{gray!10} MotionAGFormer-B & 243 & $\times$ & 11.7 M & 48.3 G & 198 M & \textbf{38.4}/32.6 & 19.4 \\
          \rowcolor{gray!10} MotionAGFormer-L & 243 & $\times$ & 19.0 M & 78.3 G & 322 M & \textbf{38.4}/\underline{32.5} & \textbf{17.3} \\
          \hline
        \end{tabular}%
    \label{tab:human3.6m-comparison}
    \end{table*}

    \subsection{Performance comparison on Human3.6M}
    We compare our MotionAGFormer with other models on Human3.6M. For a fair comparison, only results of models without extra pre-training on additional data is included.  The results, outlined in Table~\ref{tab:human3.6m-comparison}, demonstrate that MotionAGFormer-L attains a P1 error of 38.4\,mm for estimated 2D pose and 17.3\,mm for ground truth 2D pose. Remarkably, this is achieved with approximately half the computational requirements and parameters compared to the previous SOTA (MotionBERT~\cite{motionbert}), while being 0.8\,mm and 0.5\,mm more accurate, respectively. When comparing the Base and Large variants, MotionAGFormer-L shares the same P1 error as MotionAGFormer-B in the presence of noisy data, while exhibiting a 2.1\,mm reduction in error when using ground truth 2D data. Furthermore, MotionAGFormer-S takes in only a third of the frames compared to the baselines, yet it manages to attain a superior P1 error compared to a majority of them. Similarly, MotionAGFormer-XS delivers comparable performance to PoseTransformerV2, even though it receives only a ninth of the input information (27 vs. 243 frames) and operates with approximately seven times fewer parameters.
    
    \subsection{Performance comparison on MPI-INF-3DHP}
    In evaluating our method on the MPI-INF-3DHP dataset, we modified our base and large variants to use 81 frames due to shorter video sequences. Across all variants, our method consistently outperforms others in terms MPJPE. Notably, our large variant achieves remarkable results with an 85.3\% AUC and a 16.2\,mm P1 error. This outperforms the best models by a significant margin of 1.4\% in AUC and 6.9\,mm in P1 error. However, it achieves 98.2\% PCK, which is 0.5\% lower than the PCK performance of the compared models (Table~\ref{tab:MPI-INF-3DHP-comparison}).
    \begin{table}[h]\small
    \caption{Quantitative comparisons on MPI-INF-3DHP. $T$: Number of input frames. The best and second-best scores are in bold and underlined, respectively.}
      \centering
      \resizebox{1\linewidth}{!}{%
        \begin{tabular}{lc|ccc}
          \hline
          Method & $T$ & PCK$\uparrow$ & AUC$\uparrow$ & MPJPE$\downarrow$ \\
          \hline
          MHFormer~\cite{li2022mhformer} & 9 & 93.8 & 63.3 & 58.0 \\
          MixSTE~\cite{mixste} & 27 & 94.4 & 66.5 & 54.9 \\
          P-STMO~\cite{pstmo} & 81 & 97.9 & 75.8 & 32.2 \\ 
          Einfalt \textit{et al.}~\cite{einfalt_up3dhpe_WACV23} & 81 & 95.4 & 67.6 & 46.9 \\
          STCFormer~\cite{STCFormer} & 81 & \textbf{98.7} & 83.9 & 23.1 \\
          PoseFormerV2~\cite{poseformerv2} & 81 & 97.9 & 78.8 & 27.8 \\
          GLA-GCN~\cite{yu2023glagcn} & 81 & \underline{98.5} & 79.1 & 27.7 \\
          HSTFormer~\cite{qian2023hstformer} & 81 & 97.3 & 71.5 & 41.4 \\
          HDFormer~\cite{chen2023hdformer} & 96 & \textbf{98.7} & 72.9 & 37.2 \\
          \rowcolor{gray!10} MotionAGFormer-XS & 27 & 98.2 & 83.5 & 19.2 \\
          \rowcolor{gray!10} MotionAGFormer-S & 81 & 98.3 & \underline{84.5} & \underline{17.1} \\
          \rowcolor{gray!10} MotionAGFormer-B & 81 & 98.3 & 84.2 & 18.2 \\
          \rowcolor{gray!10} MotionAGFormer-L & 81 & 98.2 & \textbf{85.3} & \textbf{16.2} \\
          \hline
        \end{tabular}%
      }
    \label{tab:MPI-INF-3DHP-comparison}
    \end{table}

    \subsection{Ablation Studies}
    \label{sec:ablation}
    A series of ablation studies were conducted on the Human3.6M dataset to explore different design choices for the AGFormer block.

    The initial ablation study investigates the influence of favoring the number of AGFormer blocks and the width of our model on the P1 error. As shown in Table~\ref{tab:width-depth-comparison}, the trend generally leans towards favoring a model that is deeper but narrower. Interestingly, a model with 16 AGFormer blocks and a width of 128 exhibits similar performance to a model featuring 12 AGFormer blocks with a width of 256, while the first configuration uses approximately three times less memory and computational resources.
    
    The second part of the ablation study explores alternative modules for the graph stream within the AGFormer block. The results are presented in Table~\ref{tab:graph-comparison}. Shifting from GCNFormer to GCN blocks degrades the P1 error by 0.7\,mm. Similarly, substituting the Temporal GCNFormer with TCN degrades the P1 error by 0.2\,mm when using GCN and 0.9\,mm when using GCNFormer. Lastly, replacing the GCN with a CTR-GCN~\cite{ctrn-gcn} increases the P1 error by 2.5\,mm. We hypothesize that is because of the continuous refinement of topology for each channel in CTR-GCN during training. As a result, every joint not only obtains information from its immediate neighbors but also from all other joints. This prevents the generation of complementary information for the transformer stream.

\begin{figure}[t]
  \centering
  \includegraphics[width=\linewidth]{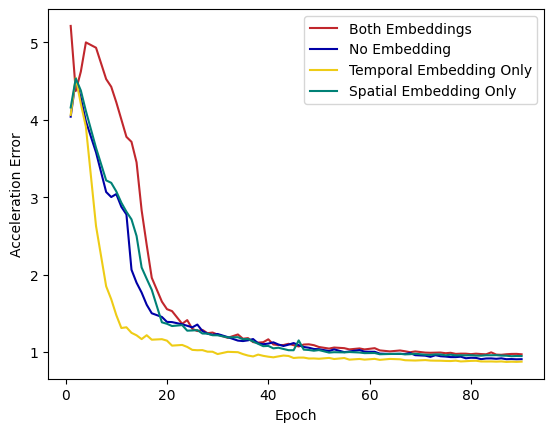}
  \caption{\textbf{Acceleration error when using different type of positional embedding.} Using temporal positional embedding leads to 0.88\,mm error and a faster convergence compared to spatial positional embedding, without embedding, and both embeddings that reach 0.94\,mm, 0.90\,mm, and 0.97\,mm respectively.}
  \label{fig:acc-error}
\end{figure}

    To explore the impact of positional embedding on the the final performance, we conducted a series of experiments outlined in Table~\ref{tab:embed-comparison}. Surprisingly, including temporal positional embedding in addition to spatial positional embedding leads to a 0.6\,mm increase in P1 error. As mentioned in Section~\ref{sec:overall-architecture}, this result stems from the non-permutation equivariant nature of the GCNFormer stream. Unlike transformers, our network inherently maintains the temporal sequence of frames. However, as shown in Figure~\ref{fig:acc-error}, adding temporal positional embedding leads to a better acceleration error (0.88\,mm) compared to using spatial positional embedding (0.94\,mm).

    Finally, to verify the efficiency of the proposed AGFormer block, Table~\ref{tab:metaformer-comparison} shows alternative blocks. When using GCNFormer, a P1 error of 57.5\,mm is observed, indicating its limited capability to accurately capture the underlying 3D sequence structure. Nevertheless, a hybrid approach involving both GCNFormer and Transformer yields a noteworthy improvement, reducing the P1 error by 5.2\,mm compared to using Transformer alone. Furthermore, the sequential fusion of these two modules is not as effective as their parallel integration.

    \begin{table}[h]\small
    \caption{The P1 error comparison by varying number of AGFormer blocks and number of channels. $d$: Number of channels in each AGFormer block. $d'$: Number of channels before regression head. $T$ is kept 243 in all experiments.}
      \centering
        \begin{tabular}{cccccc}
          \hline
          Layers & $d$ & $d'$ & Param & MACs & P1 \\
          \hline
          5 & 512 & 512 & 58 M & 252.8 G & 39.2 \\
          4 & 256 & 512 & 11.7 M & 54.G G & 40.2 \\
          6 & 256 & 512 & 17.5 M & 81.64 G & 39.6 \\
          8 & 256 & 512 & 23.3 M & 108.6 G & 39.5 \\
          10 & 256 & 512 & 29.1 M & 135.7 G & 38.6 \\
          12 & 256 & 512 & 34.9 M & 162.7 G & 38.4 \\
          14 & 256 & 512 & 40.7 M & 189.7 G & 38.6 \\
          16 & 128 & 512 & 11.7 M & 64.7 G & 38.4 \\
          16 & 128 & 256 & 11.7 M & 64.5 G & 38.7 \\
          16 & 128 & 1024 & 11.7 M & 64.9 G & 39.2 \\
          26 & 64 & 512 & 4.8 M & 39.5 G & 39.5 \\
          \hline
        \end{tabular}%
    \label{tab:width-depth-comparison}
    \end{table}

    \begin{table}[h]\small
    \caption{The P1 error comparison when using alternative modules in graph stream of AGFormer block.}
      \centering
        \begin{tabular}{cccc|c}
          \hline
          GCN & TCN & GCNFormer & CTR-GCNFormer & P1 \\
          \hline
          \checkmark & - & - & - & 39.1 \\
          \checkmark & \checkmark & - & - & 39.3 \\
          - & \checkmark & \checkmark & - & 39.7 \\
          - & - & - & \checkmark & 40.9 \\
          - & - & \checkmark & - & 38.4 \\
          \hline
        \end{tabular}%
    \label{tab:graph-comparison}
    \end{table}
    
    \subsection{Qualitative Analysis}
    Validation of MotionAGFormer-B is conducted using the adjacency matrix of temporal GCNFormer and visualization of 3D human pose estimation. The instances for validation are randomly chosen from the evaluation set of Human3.6M.

   \textbf{Qualitative Comparisons.} Figure~\ref{fig:qualitative-comparison} compares MotionAGFormer-B with recent approaches including STCFormer~\cite{STCFormer}, PoseFormerV2~\cite{poseformerv2}, and MotionBERT~\cite{motionbert}. By design, confidence scores of the 2D detector are only included into MotionBERT and MotionAGFormer-B. Overall, MotionAGFormer-B shows better reconstruction results across all three samples than PoseFormerV2 and STCFormer, while maintaining competitive performance with MotionBERT. Specifically, in Figure~\ref{fig:qualitative-comparison}a, MotionAGFormer exhibits improved alignment with the ground truth in comparison to alternative approaches. In Figure~\ref{fig:qualitative-comparison}b, it displays a slightly superior alignment, whereas in Figure~\ref{fig:qualitative-comparison}c, its alignment is marginally less optimal than that of MotionBERT.
   
\begin{figure}[!t]
  \centering
  \includegraphics[width=\linewidth]{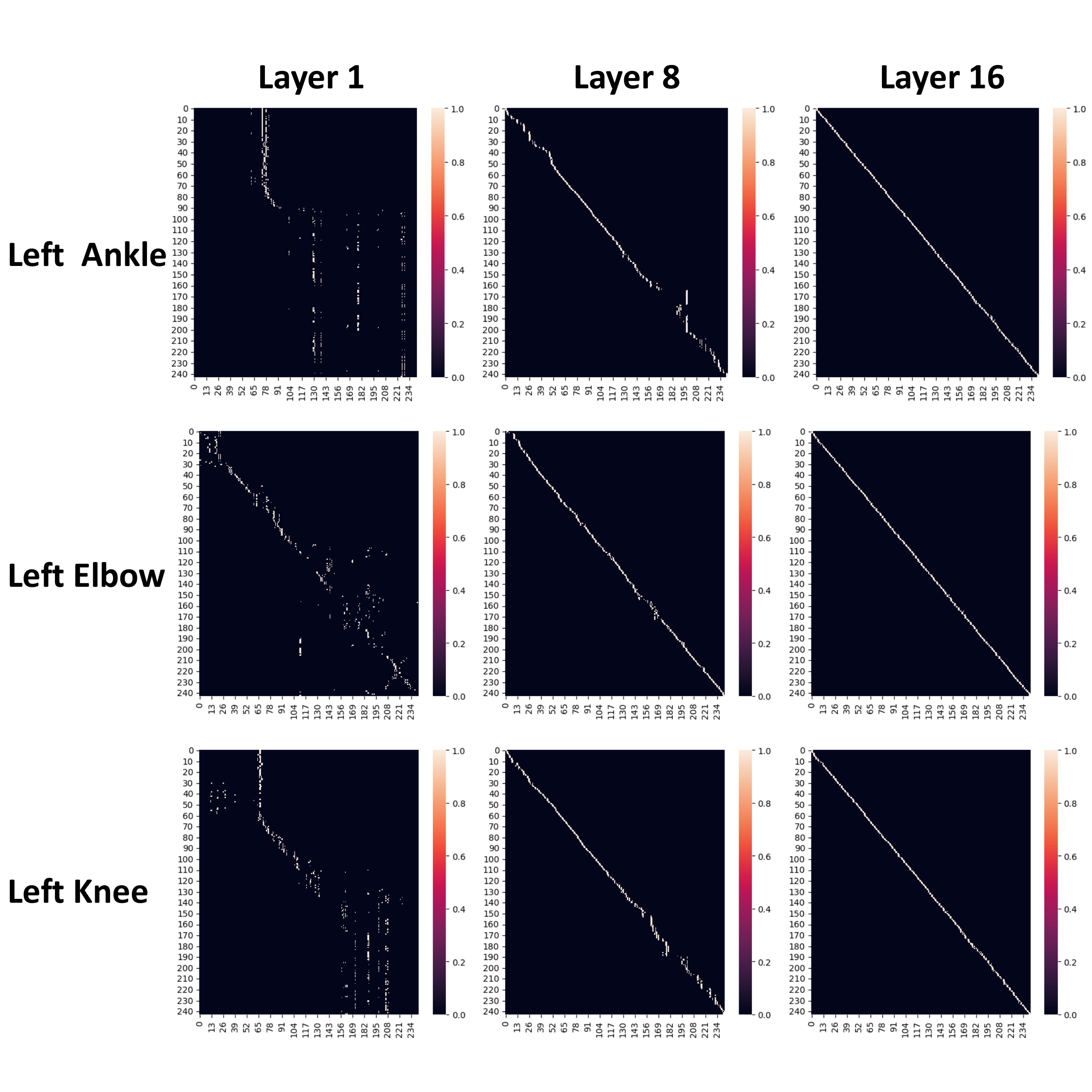}
  \caption{Temporal adjacency matrix of a random sequence on three different joints, from the Human3.6M dataset, at the first layer (left), middle layer (center), and last layer (right). $K$ in K-NN is set to 2.}
  \label{fig:temporal-adj}
\end{figure}

   \textbf{Temporal adjacency visualization.} The adjacency matrix of temporal GCNFormer is visualized in Figure~\ref{fig:temporal-adj}. In the initial layers, individual joints exhibit distinct adjacency matrices, which should vary across different sequences due to the changing joint positions over time. Nevertheless, as we progress through the model's depth, it appears to learn a representation where each joint is most akin to its own state in neighboring frames. Consequently, this leads to connections being formed with adjacent frames.

    \begin{table}[h]\small
    \caption{The P1 error comparison when using different positional embedding.}
      \centering
        \begin{tabular}{cc|c}
          \hline
          Temporal Embedding & Spatial Embedding & P1 \\
          \hline
          - & - & 39.3 \\
          -  & \checkmark & 38.4 \\
          \checkmark  & - & 38.9 \\
          \checkmark & \checkmark & 40.5 \\
          \hline
        \end{tabular}%
    \label{tab:embed-comparison}
    \end{table}

    \begin{table}[h]
    \caption{\textbf{Comparison of different MetaFormer integration.} All the models are trained on Human3.6M with our MotionAGFormer-B settings.}
      \centering
        \begin{tabular}{lc}
          \hline
          Method & P1 \\
          \hline
          GCNFormer only & 57.5\\
          Transformer only & 43.6\\
          GCNFormer~$\rightarrow~$Transformer~(Sequential) & 39.1\\
          Transformer~$\rightarrow$~GCNFormer~(Sequential) & 38.9\\
          Transformer~-~GCNFormer~(Parallel) & 38.4\\
          \hline
        \end{tabular}%
    \label{tab:metaformer-comparison}
    \end{table}

\section{Conclusion}
We introduced MotionAGFormer, a novel approach that leverages GCNFormer to capture intricate local joint relationships, and combines it with Transformer that effectively captures global joint interdependencies. This fusion enhances the model's ability to comprehend the inherent 3D structure within input 2D sequences. Additionally, MotionAGFormer offers various adaptable variants, allowing the selection of an optimal balance between speed and accuracy. Empirical evaluations show that our method surpasses alternative methods on Human3.6M and MPI-INF-3DHP.

{\small
\bibliographystyle{ieee_fullname}
\bibliography{egbib}

\begin{thebibliography}{10}\itemsep=-1pt

\bibitem{ba2016layer}
Jimmy~Lei Ba, Jamie~Ryan Kiros, and Geoffrey~E Hinton.
\newblock Layer normalization.
\newblock {\em arXiv preprint arXiv:1607.06450}, 2016.

\bibitem{bauer2023weakly}
Peter Bauer, Arij Bouazizi, Ulrich Kressel, and Fabian~B Flohr.
\newblock Weakly supervised multi-modal 3d human body pose estimation for autonomous driving.
\newblock In {\em 2023 IEEE Intelligent Vehicles Symposium (IV)}, pages 1--7. IEEE, 2023.

\bibitem{chen2023hdformer}
Hanyuan Chen, Jun-Yan He, Wangmeng Xiang, Wei Liu, Zhi-Qi Cheng, Hanbing Liu, Bin Luo, Yifeng Geng, and Xuansong Xie.
\newblock {HDFormer}: High-order directed transformer for 3d human pose estimation.
\newblock {\em arXiv preprint arXiv:2302.01825}, 2023.

\bibitem{cpn}
Yilun Chen, Zhicheng Wang, Yuxiang Peng, Zhiqiang Zhang, Gang Yu, and Jian Sun.
\newblock Cascaded pyramid network for multi-person pose estimation.
\newblock In {\em Proceedings of the IEEE conference on computer vision and pattern recognition}, pages 7103--7112, 2018.

\bibitem{ctrn-gcn}
Yuxin Chen, Ziqi Zhang, Chunfeng Yuan, Bing Li, Ying Deng, and Weiming Hu.
\newblock Channel-wise topology refinement graph convolution for skeleton-based action recognition.
\newblock In {\em Proceedings of the IEEE/CVF international conference on computer vision}, pages 13359--13368, 2021.

\bibitem{choi2020pose2mesh}
Hongsuk Choi, Gyeongsik Moon, and Kyoung~Mu Lee.
\newblock {Pose2Mesh}: Graph convolutional network for 3d human pose and mesh recovery from a 2d human pose.
\newblock In {\em Computer Vision--ECCV 2020: 16th European Conference, Glasgow, UK, August 23--28, 2020, Proceedings, Part VII 16}, pages 769--787. Springer, 2020.

\bibitem{chun2023learnable}
Sungho Chun, Sungbum Park, and Ju~Yong Chang.
\newblock Learnable human mesh triangulation for 3d human pose and shape estimation.
\newblock In {\em Proceedings of the IEEE/CVF Winter Conference on Applications of Computer Vision}, pages 2850--2859, 2023.

\bibitem{ci2019optimizing}
Hai Ci, Chunyu Wang, Xiaoxuan Ma, and Yizhou Wang.
\newblock Optimizing network structure for 3d human pose estimation.
\newblock In {\em Proceedings of the IEEE/CVF international conference on computer vision}, pages 2262--2271, 2019.

\bibitem{czech2022board}
Phillip Czech, Markus Braun, Ulrich Kre{\ss}el, and Bin Yang.
\newblock On-board pedestrian trajectory prediction using behavioral features.
\newblock In {\em 2022 21st IEEE International Conference on Machine Learning and Applications (ICMLA)}, pages 437--443. IEEE, 2022.

\bibitem{ViT}
Alexey Dosovitskiy, Lucas Beyer, Alexander Kolesnikov, Dirk Weissenborn, Xiaohua Zhai, Thomas Unterthiner, Mostafa Dehghani, Matthias Minderer, Georg Heigold, Sylvain Gelly, et~al.
\newblock An image is worth 16x16 words: Transformers for image recognition at scale.
\newblock {\em arXiv preprint arXiv:2010.11929}, 2020.

\bibitem{einfalt_up3dhpe_WACV23}
Moritz Einfalt, Katja Ludwig, and Rainer Lienhart.
\newblock Uplift and upsample: Efficient 3d human pose estimation with uplifting transformers.
\newblock In {\em Proceedings of the IEEE/CVF Winter Conference on Applications of Computer Vision (WACV)}, January 2023.

\bibitem{foo2023unified}
Lin~Geng Foo, Tianjiao Li, Hossein Rahmani, Qiuhong Ke, and Jun Liu.
\newblock Unified pose sequence modeling.
\newblock In {\em Proceedings of the IEEE/CVF Conference on Computer Vision and Pattern Recognition}, pages 13019--13030, 2023.

\bibitem{ioffe2015batch}
Sergey Ioffe and Christian Szegedy.
\newblock Batch normalization: Accelerating deep network training by reducing internal covariate shift.
\newblock In {\em International conference on machine learning}, pages 448--456. pmlr, 2015.

\bibitem{ionescu2013human3}
Catalin Ionescu, Dragos Papava, Vlad Olaru, and Cristian Sminchisescu.
\newblock Human3. 6m: Large scale datasets and predictive methods for 3d human sensing in natural environments.
\newblock {\em IEEE transactions on pattern analysis and machine intelligence}, 36(7):1325--1339, 2013.

\bibitem{iskakov2019learnable}
Karim Iskakov, Egor Burkov, Victor Lempitsky, and Yury Malkov.
\newblock Learnable triangulation of human pose.
\newblock In {\em Proceedings of the IEEE/CVF international conference on computer vision}, pages 7718--7727, 2019.

\bibitem{dc-gct}
Hongbo Kang, Yong Wang, Mengyuan Liu, Doudou Wu, Peng Liu, and Wenming Yang.
\newblock Double-chain constraints for 3d human pose estimation in images and videos.
\newblock {\em arXiv preprint arXiv:2308.05298}, 2023.

\bibitem{kipf2016gcn}
Thomas~N Kipf and Max Welling.
\newblock Semi-supervised classification with graph convolutional networks.
\newblock {\em arXiv preprint arXiv:1609.02907}, 2016.

\bibitem{hdgcn}
Jungho Lee, Minhyeok Lee, Dogyoon Lee, and Sangyoun Lee.
\newblock Hierarchically decomposed graph convolutional networks for skeleton-based action recognition.
\newblock {\em arXiv preprint arXiv:2208.10741}, 2022.

\bibitem{li2022exploiting}
Wenhao Li, Hong Liu, Runwei Ding, Mengyuan Liu, Pichao Wang, and Wenming Yang.
\newblock Exploiting temporal contexts with strided transformer for 3d human pose estimation.
\newblock {\em IEEE Transactions on Multimedia}, 25:1282--1293, 2022.

\bibitem{li2022mhformer}
Wenhao Li, Hong Liu, Hao Tang, Pichao Wang, and Luc Van~Gool.
\newblock {MHFormer}: Multi-hypothesis transformer for 3d human pose estimation.
\newblock In {\em Proceedings of the IEEE/CVF Conference on Computer Vision and Pattern Recognition (CVPR)}, pages 13147--13156, 2022.

\bibitem{lin2010augmented}
Huei-Yung Lin and Ting-Wen Chen.
\newblock Augmented reality with human body interaction based on monocular 3d pose estimation.
\newblock In {\em International Conference on Advanced Concepts for Intelligent Vision Systems}, pages 321--331. Springer, 2010.

\bibitem{msg3d}
Ziyu Liu, Hongwen Zhang, Zhenghao Chen, Zhiyong Wang, and Wanli Ouyang.
\newblock Disentangling and unifying graph convolutions for skeleton-based action recognition.
\newblock In {\em Proceedings of the IEEE/CVF conference on computer vision and pattern recognition}, pages 143--152, 2020.

\bibitem{AdamW}
Ilya Loshchilov and Frank Hutter.
\newblock Decoupled weight decay regularization.
\newblock {\em arXiv preprint arXiv:1711.05101}, 2017.

\bibitem{luo2022learning}
Cheng Luo, Siyang Song, Weicheng Xie, Linlin Shen, and Hatice Gunes.
\newblock Learning multi-dimensional edge feature-based au relation graph for facial action unit recognition.
\newblock {\em arXiv preprint arXiv:2205.01782}, 2022.

\bibitem{mehta2017monocular}
Dushyant Mehta, Helge Rhodin, Dan Casas, Pascal Fua, Oleksandr Sotnychenko, Weipeng Xu, and Christian Theobalt.
\newblock Monocular 3d human pose estimation in the wild using improved cnn supervision.
\newblock In {\em 2017 international conference on 3D vision (3DV)}, pages 506--516. IEEE, 2017.

\bibitem{mehta2017vnect}
Dushyant Mehta, Srinath Sridhar, Oleksandr Sotnychenko, Helge Rhodin, Mohammad Shafiei, Hans-Peter Seidel, Weipeng Xu, Dan Casas, and Christian Theobalt.
\newblock {VNect}: Real-time 3d human pose estimation with a single rgb camera.
\newblock {\em Acm transactions on graphics (tog)}, 36(4):1--14, 2017.

\bibitem{munea2020progress}
Tewodros~Legesse Munea, Yalew~Zelalem Jembre, Halefom~Tekle Weldegebriel, Longbiao Chen, Chenxi Huang, and Chenhui Yang.
\newblock The progress of human pose estimation: A survey and taxonomy of models applied in 2d human pose estimation.
\newblock {\em IEEE Access}, 8:133330--133348, 2020.

\bibitem{stackedhourglass}
Alejandro Newell, Kaiyu Yang, and Jia Deng.
\newblock Stacked hourglass networks for human pose estimation.
\newblock In {\em Computer Vision--ECCV 2016: 14th European Conference, Amsterdam, The Netherlands, October 11-14, 2016, Proceedings, Part VIII 14}, pages 483--499. Springer, 2016.

\bibitem{newell2016stacked}
Alejandro Newell, Kaiyu Yang, and Jia Deng.
\newblock Stacked hourglass networks for human pose estimation.
\newblock In {\em Computer Vision--ECCV 2016: 14th European Conference, Amsterdam, The Netherlands, October 11-14, 2016, Proceedings, Part VIII 14}, pages 483--499. Springer, 2016.

\bibitem{pytorch}
Adam Paszke, Sam Gross, Soumith Chintala, Gregory Chanan, Edward Yang, Zachary DeVito, Zeming Lin, Alban Desmaison, Luca Antiga, and Adam Lerer.
\newblock Automatic differentiation in pytorch.
\newblock 2017.

\bibitem{pavlakos2018ordinal}
Georgios Pavlakos, Xiaowei Zhou, and Kostas Daniilidis.
\newblock Ordinal depth supervision for 3d human pose estimation.
\newblock In {\em Proceedings of the IEEE conference on computer vision and pattern recognition}, pages 7307--7316, 2018.

\bibitem{pavlakos2017coarse}
Georgios Pavlakos, Xiaowei Zhou, Konstantinos~G Derpanis, and Kostas Daniilidis.
\newblock Coarse-to-fine volumetric prediction for single-image 3d human pose.
\newblock In {\em Proceedings of the IEEE conference on computer vision and pattern recognition}, pages 7025--7034, 2017.

\bibitem{qian2023hstformer}
Xiaoye Qian, Youbao Tang, Ning Zhang, Mei Han, Jing Xiao, Ming-Chun Huang, and Ruei-Sung Lin.
\newblock {HSTFormer}: Hierarchical spatial-temporal transformers for 3d human pose estimation.
\newblock {\em arXiv preprint arXiv:2301.07322}, 2023.

\bibitem{reddy2021tessetrack}
N~Dinesh Reddy, Laurent Guigues, Leonid Pishchulin, Jayan Eledath, and Srinivasa~G Narasimhan.
\newblock {TesseTrack}: End-to-end learnable multi-person articulated 3d pose tracking.
\newblock In {\em Proceedings of the IEEE/CVF Conference on Computer Vision and Pattern Recognition}, pages 15190--15200, 2021.

\bibitem{remelli2020lightweight}
Edoardo Remelli, Shangchen Han, Sina Honari, Pascal Fua, and Robert Wang.
\newblock Lightweight multi-view 3d pose estimation through camera-disentangled representation.
\newblock In {\em Proceedings of the IEEE/CVF conference on computer vision and pattern recognition}, pages 6040--6049, 2020.

\bibitem{pstmo}
Wenkang Shan, Zhenhua Liu, Xinfeng Zhang, Shanshe Wang, Siwei Ma, and Wen Gao.
\newblock {P-STMO}: Pre-trained spatial temporal many-to-one model for 3d human pose estimation.
\newblock In {\em Computer Vision--ECCV 2022: 17th European Conference, Tel Aviv, Israel, October 23--27, 2022, Proceedings, Part V}, pages 461--478. Springer, 2022.

\bibitem{su2023towards}
Weijie Su, Xizhou Zhu, Chenxin Tao, Lewei Lu, Bin Li, Gao Huang, Yu Qiao, Xiaogang Wang, Jie Zhou, and Jifeng Dai.
\newblock Towards all-in-one pre-training via maximizing multi-modal mutual information.
\newblock In {\em Proceedings of the IEEE/CVF Conference on Computer Vision and Pattern Recognition}, pages 15888--15899, 2023.

\bibitem{hrnet}
Ke Sun, Bin Xiao, Dong Liu, and Jingdong Wang.
\newblock Deep high-resolution representation learning for human pose estimation.
\newblock In {\em Proceedings of the IEEE/CVF conference on computer vision and pattern recognition}, pages 5693--5703, 2019.

\bibitem{sun2018integral}
Xiao Sun, Bin Xiao, Fangyin Wei, Shuang Liang, and Yichen Wei.
\newblock Integral human pose regression.
\newblock In {\em Proceedings of the European conference on computer vision (ECCV)}, pages 529--545, 2018.

\bibitem{STCFormer}
Zhenhua Tang, Zhaofan Qiu, Yanbin Hao, Richang Hong, and Ting Yao.
\newblock {3D} human pose estimation with spatio-temporal criss-cross attention.
\newblock In {\em Proceedings of the IEEE/CVF Conference on Computer Vision and Pattern Recognition}, pages 4790--4799, 2023.

\bibitem{vaswani2017attention}
Ashish Vaswani, Noam Shazeer, Niki Parmar, Jakob Uszkoreit, Llion Jones, Aidan~N Gomez, {\L}ukasz Kaiser, and Illia Polosukhin.
\newblock Attention is all you need.
\newblock {\em Advances in neural information processing systems}, 30, 2017.

\bibitem{wang2020motion}
Jingbo Wang, Sijie Yan, Yuanjun Xiong, and Dahua Lin.
\newblock Motion guided 3d pose estimation from videos.
\newblock In {\em European Conference on Computer Vision}, pages 764--780. Springer, 2020.

\bibitem{wiederer2020traffic}
Julian Wiederer, Arij Bouazizi, Ulrich Kressel, and Vasileios Belagiannis.
\newblock Traffic control gesture recognition for autonomous vehicles.
\newblock In {\em 2020 IEEE/RSJ International Conference on Intelligent Robots and Systems (IROS)}, pages 10676--10683. IEEE, 2020.

\bibitem{yu2023glagcn}
Bruce X.~B. Yu, Zhi Zhang, Yongxu Liu, Sheng hua Zhong, Yan Liu, and Chang~Wen Chen.
\newblock {GLA-GCN}: Global-local adaptive graph convolutional network for 3d human pose estimation from monocular video, 2023.

\bibitem{yu2022coca}
Jiahui Yu, Zirui Wang, Vijay Vasudevan, Legg Yeung, Mojtaba Seyedhosseini, and Yonghui Wu.
\newblock {CoCa}: Contrastive captioners are image-text foundation models.
\newblock {\em arXiv preprint arXiv:2205.01917}, 2022.

\bibitem{yu2022metaformer}
Weihao Yu, Mi Luo, Pan Zhou, Chenyang Si, Yichen Zhou, Xinchao Wang, Jiashi Feng, and Shuicheng Yan.
\newblock {MetaFormer} is actually what you need for vision.
\newblock In {\em Proceedings of the IEEE/CVF Conference on Computer Vision and Pattern Recognition}, pages 10819--10829, 2022.

\bibitem{mixste}
Jinlu Zhang, Zhigang Tu, Jianyu Yang, Yujin Chen, and Junsong Yuan.
\newblock {MixSTE}: Seq2seq mixed spatio-temporal encoder for 3d human pose estimation in video.
\newblock In {\em Proceedings of the IEEE/CVF Conference on Computer Vision and Pattern Recognition (CVPR)}, pages 13232--13242, June 2022.

\bibitem{zhang2021adafuse}
Zhe Zhang, Chunyu Wang, Weichao Qiu, Wenhu Qin, and Wenjun Zeng.
\newblock {Adafuse}: Adaptive multiview fusion for accurate human pose estimation in the wild.
\newblock {\em International Journal of Computer Vision}, 129:703--718, 2021.

\bibitem{zhao2019semantic}
Long Zhao, Xi Peng, Yu Tian, Mubbasir Kapadia, and Dimitris~N Metaxas.
\newblock Semantic graph convolutional networks for 3d human pose regression.
\newblock In {\em Proceedings of the IEEE/CVF conference on computer vision and pattern recognition}, pages 3425--3435, 2019.

\bibitem{poseformerv2}
Qitao Zhao, Ce Zheng, Mengyuan Liu, Pichao Wang, and Chen Chen.
\newblock {PoseFormerV2}: Exploring frequency domain for efficient and robust 3d human pose estimation.
\newblock In {\em Proceedings of the IEEE/CVF Conference on Computer Vision and Pattern Recognition (CVPR)}, pages 8877--8886, June 2023.

\bibitem{poseformer}
Ce Zheng, Sijie Zhu, Matias Mendieta, Taojiannan Yang, Chen Chen, and Zhengming Ding.
\newblock 3d human pose estimation with spatial and temporal transformers.
\newblock In {\em Proceedings of the IEEE/CVF International Conference on Computer Vision}, pages 11656--11665, 2021.

\bibitem{zhou2019hemlets}
Kun Zhou, Xiaoguang Han, Nianjuan Jiang, Kui Jia, and Jiangbo Lu.
\newblock {HEMlets} pose: Learning part-centric heatmap triplets for accurate 3d human pose estimation.
\newblock In {\em Proceedings of the IEEE/CVF international conference on computer vision}, pages 2344--2353, 2019.

\bibitem{motionbert}
Wentao Zhu, Xiaoxuan Ma, Zhaoyang Liu, Libin Liu, Wayne Wu, and Yizhou Wang.
\newblock {MotionBERT}: A unified perspective on learning human motion representations.
\newblock In {\em Proceedings of the IEEE/CVF International Conference on Computer Vision}, 2023.

\bibitem{zong2022detrs}
Zhuofan Zong, Guanglu Song, and Yu Liu.
\newblock {DETRs} with collaborative hybrid assignments training.
\newblock {\em arXiv preprint arXiv:2211.12860}, 2022.

\end{thebibliography}
}
\appendix
\section*{Appendix}
\begin{figure}[!t]
          \centering
          \includegraphics[width=0.5\textwidth]{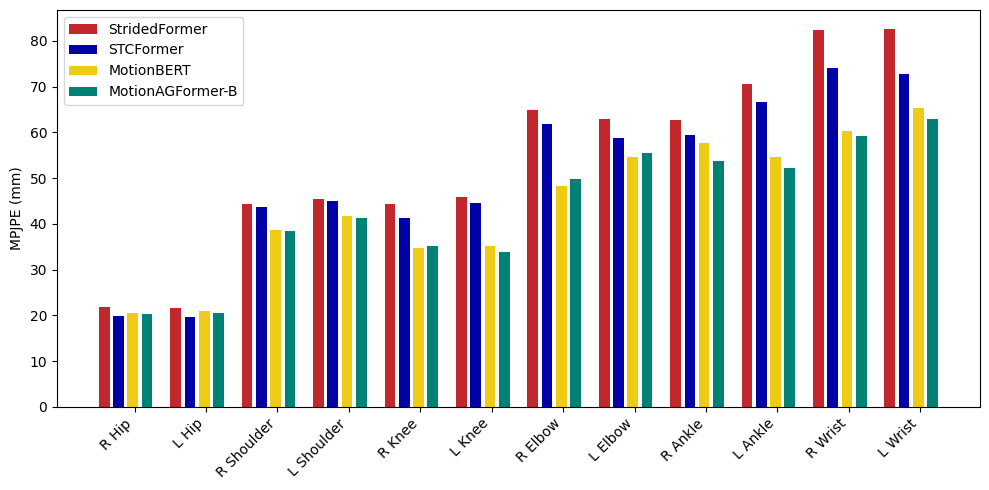}
          \caption{The per-joint error comparisons in terms of MPJPE (mm) with other models on Human3.6M dataset. 'L' and 'R' denote left and right, respectively.}
          \label{fig:per-joint-error}
    \end{figure}
    \section{Additional Results on Human3.6M}
    \subsection{Per-action Result}
    \begin{table*}[h]\small
    \caption{Quantitative comparisons of 3D human pose estimation per action on Human3.6M. (Top) MPJPE~(mm) using detected 2D pose sequence. (Bottom) P-MPJPE~(mm) using detected 2D pose sequence. (*) denotes using HRNet~\cite{hrnet} for 2D pose estimation. (\dag) denotes manually evaluated using their provided evaluation code.}
      \centering
      \resizebox{\linewidth}{!}{
        \begin{tabular}{lc|ccccccccccccccc|c}
          \hline
          \textbf{MPJPE} & $T$ & Dire. & Disc. & Eat & Greet & Phone & Photo & Pose & Purch. & Sit & SitD & Smoke & Wait & WalkD & Walk & WalkT & Avg\\
          \hline
          *MHFormer~\cite{li2022mhformer} & 351 & 39.2 & 43.1 & 40.1 & 40.9 & 44.9 & 51.2 & 40.6 & 41.3 & 53.5 & 60.3 & 43.7 & 41.1 & 43.8 & 29.8 & 30.6 & 43.0\\
          MixSTE~\cite{mixste} & 243 & 37.6 & 40.9 & 37.3 & 39.7 & 42.3 & 49.9 & 40.1 & 39.8 & 51.7 & 55.0 & 42.1 & 39.8 & 41.0 & 27.9 & 27.9 & 40.9\\
          P-STMO~\cite{pstmo} & 243 & 38.9 & 42.7 & 40.4 & 41.1 & 45.6 & 49.7 & 40.9 & 39.9 & 55.5 & 59.4 & 44.9 & 42.2 & 42.7 & 29.4 & 29.4 & 42.8\\
          StridedFormer~\cite{li2022exploiting} & 351 & 40.3 & 43.3 & 40.2 & 42.3 & 45.6 & 52.3 & 41.8 & 40.5 & 55.9 & 60.6 & 44.2 & 43.0 & 44.2 & 30.0 & 30.2 & 43.7\\
          Einfalt~\textit{et al.}~\cite{einfalt_up3dhpe_WACV23} & 351 & 39.6 & 43.8 & 40.2 & 42.4 & 46.5 & 53.9 & 42.3 & 42.5 & 55.7 & 62.3 & 45.1 & 43.0 & 44.7 & 30.1 & 30.8 & 44.2\\
          STCFormer~\cite{STCFormer} & 243 & 39.6 & 41.6 & 37.4 & 38.8 & 43.1 & 51.1 & 39.1 & 39.7 & 51.4 & 57.4 & 41.8 & 38.5 & 40.7 & 27.1 & 28.6 & 41.0\\
          STCFormer-L~\cite{STCFormer} & 243 & 38.4 & 41.2 & \underline{36.8} & 38.0 & 42.7 & 50.5 & 38.7 & 38.2 & 52.5 & 56.8 & 41.8 & 38.4 & 40.2 & \textbf{26.2} & 27.7 & 40.5\\
          UPS~\cite{foo2023unified} & 243 & 37.5 & 39.2 & 36.9 & 40.6 & \textbf{39.3} & \textbf{46.8} & 39.0 & 41.7 & 50.6 & 63.5 & \underline{40.4} & 37.8 & 44.2 & \underline{26.7} & 29.1 & 40.8\\
          GLA-GCN~\cite{yu2023glagcn} & 243 & 41.3 & 44.3 & 40.8 & 41.8 & 45.9 & 54.1 & 42.1 & 41.5 & 57.8 & 62.9 & 45.0 & 42.8 & 45.9 & 29.4 & 29.9 & 44.4\\
          \dag~MotionBERT~\cite{motionbert} & 243 & \underline{36.6} & 39.3 & 37.8 & 33.5 & 41.4 & 49.9 & \textbf{37.0} & 35.5 & \underline{50.4} & 56.5 & 41.4 & 38.2 & 37.3 & \textbf{26.2} & \textbf{26.9} & \underline{39.2}\\
          HDFormer~\cite{chen2023hdformer} & 96 & 38.1 & 43.1 & 39.3 & 39.4 & 44.3 & 49.1 & 41.3 & 40.8 & 53.1 & 62.1 & 43.3 & 41.8 & 43.1 & 31.0 & 29.7 & 42.6\\
          HSTFormer~\cite{qian2023hstformer} & 81 & 39.5 & 42.0 & 39.9 & 40.8 & 44.4 & 50.9 & 40.9 & 41.3 & 54.7 & 58.8 & 43.6 & 40.7 & 43.4 & 30.1 & 30.4 & 42.7\\
          \rowcolor{gray!10} MotionAGFormer-XS & 27 & 42.7 & 46.7 & 41.1 & 39.4 & 47.0 & 56.3 & 44.4 & 41.0 & 52.4 & 66.3 & 46.7 & 43.2 & 43.4 & 32.4 & 34.0 & 45.1\\
          \rowcolor{gray!10} MotionAGFormer-S & 81 & 41.9 & 42.7 & 40.4 & 37.6 & 45.6 & 51.3 & 41.0 & 38.0 & 54.1 & 58.8 & 45.5 & 40.4 & 39.8 & 29.4 & 31.0 & 42.5\\
          \rowcolor{gray!10} MotionAGFormer-B & 243 & \textbf{36.4} & \textbf{38.4} & \underline{36.8} & \textbf{32.9} & \underline{40.9} & \underline{48.5} & \underline{36.6} & \textbf{34.6} & 51.7 & \underline{52.8} & 41.0 & \textbf{36.4} & \underline{36.5} & \underline{26.7} & \underline{27.0} & \textbf{38.4}\\
          \rowcolor{gray!10} MotionAGFormer-L & 243 & 36.8 & \underline{38.5} & \textbf{35.9} & \underline{33.0} & 41.1 & 48.6 & 38.0 & \underline{34.8} & \textbf{49.0} & \textbf{51.4} & \textbf{40.3} & \underline{37.4} & \textbf{36.3} & 27.2 & 27.2 & \textbf{38.4}\\
          \hline
          \hline
          \textbf{P-MPJPE} & $T$ & Dire. & Disc. & Eat & Greet & Phone & Photo & Pose & Purch. & Sit & SitD & Smoke & Wait & WalkD & Walk & WalkT & Avg\\
          *MHFormer~\cite{li2022mhformer} & 351 & 31.5 & 34.9 & 32.8 & 33.6 & 35.3 & 39.6 & 32.0 & 32.2 & 43.5 & 48.7 & 36.4 & 32.6 & 34.3 & 23.9 & 25.1 & 34.4\\
          MixSTE~\cite{mixste} & 243 & 30.8 & 33.1 & \textbf{30.3} & 31.8 & 33.1 & 39.1 & 31.1 & 30.5 & 42.5 & \textbf{44.5} & 34.0 & 30.8 & 32.7 & 22.1 & 22.9 & 32.6\\
          P-STMO~\cite{pstmo} & 243 & 31.3 & 35.2 & 32.9 & 33.9 & 35.4 & 39.3 & 32.5 & 31.5 & 44.6 & 48.2 & 36.3 & 32.9 & 34.4 & 23.8 & 23.9 & 34.4\\
          StridedFormer~\cite{li2022exploiting} & 351 & 32.7 & 35.5 & 32.5 & 35.4 & 35.9 & 41.6 & 33.0 & 31.9 & 45.1 & 50.1 & 36.3 & 33.5 & 35.1 & 23.9 & 25.0 & 35.2\\
          Einfalt~\textit{et al.}~\cite{einfalt_up3dhpe_WACV23} & 351 & 32.7 & 36.1 & 33.4 & 36.0 & 36.1 & 42.0 & 33.3 & 33.1 & 45.4 & 50.7 & 37.0 & 34.1 & 35.9 & 24.4 & 25.4 & 35.7\\
          STCFormer~\cite{STCFormer} & 243 & \underline{29.5} & 33.2 & \underline{30.6} & 31.0 & 33.0 & 38.0 & 30.4 & \underline{29.4} & 41.8 & 45.2 & 33.6 & 29.5 & 31.6 & \underline{21.3} & \underline{22.6} & \underline{32.0}\\
          STCFormer-L~\cite{STCFormer} & 243 & \textbf{29.3} & 33.0 & 30.7 & 30.6 & \underline{32.7} & 38.2 & \textbf{29.7} & \textbf{28.8} & 42.2 & \underline{45.0} & \underline{33.3} & \underline{29.4} & 31.5 & \textbf{20.9} & \textbf{22.3} & \textbf{31.8}\\
          UPS~\cite{foo2023unified} & 243 & 30.3 & 32.2 & 30.8 & 33.1 & \textbf{31.1} & \textbf{35.2} & 30.3 & 32.1 & \textbf{39.4} & 49.6 & \textbf{32.9} & \textbf{29.2} & 33.9 & 21.6 & 24.5 & 32.5\\
          GLA-GCN~\cite{yu2023glagcn} & 243 & 32.4 & 35.3 & 32.6 & 34.2 & 35.0 & 42.1 & 32.1 & 31.9 & 45.5 & 49.5 & 36.1 & 32.4 & 35.6 & 23.5 & 24.7 & 34.8\\
          \dag~MotionBERT~\cite{motionbert} & 243 & 30.8 & \underline{32.8} & 32.4 & 28.7 & 34.3 & 38.9 & \underline{30.1} & 30.0 & 42.5 & 49.7 & 36.0 & 30.8 & 22.0 & 31.7 & 23.0 & 32.9\\
          HDFormer~\cite{chen2023hdformer} & 96 & 29.6 & 33.8 & 31.7 & 31.3 & 33.7 & \underline{37.7} & 30.6 & 31.0 & \underline{41.4} & 47.6 & 35.0 & 30.9 & 33.7 & 25.3 & 23.6 & 33.1\\
          HSTFormer~\cite{qian2023hstformer} & 81 & 31.1 & 33.7 & 33.0 & 33.2 & 33.6 & 38.8 & 31.9 & 31.5 & 43.7 & 46.3 & 35.7 & 31.5 & 33.1 & 24.2 & 24.5 & 33.7\\
          \rowcolor{gray!10} MotionAGFormer-XS & 27 & 34.4 & 37.6 & 34.7 & 33.0 & 38.4 & 43.4 & 34.7 & 33.8 & 44.6 & 53.6 & 39.4 & 34.5 & 36.2 & 26.4 & 28.5 & 36.9\\
          \rowcolor{gray!10} MotionAGFormer-S & 81 & 33.7 & 35.0 & 34.8 & 31.2 & 37.9 & 40.8 & 33.2 & 32.6 & 45.3 & 50.5 & 38.7 & 32.7 & 33.4 & 24.1 & 25.7 & 35.3\\
          \rowcolor{gray!10} MotionAGFormer-B & 243 & 30.6 & \textbf{32.6} & 32.2 & \underline{28.2} & 33.8 & 38.6 & 30.5 & 29.9 & 43.3 & 47.0 & 35.2 & 29.8 & \underline{31.4} & 22.7 & 23.5 & 32.6\\
          \rowcolor{gray!10} MotionAGFormer-L & 243 & 31.0 & \textbf{32.6} & 31.0 & \textbf{27.9} & 34.0 & 38.7 & 31.5 & 30.0 & \underline{41.4} & 45.4 & 34.8 & 30.8 & \textbf{31.3} & 22.8 & 23.2 & 32.5\\
          \hline
        \end{tabular}
    }
    \label{tab:human3.6m-comparison-action}
    \end{table*}
    To assess the effectiveness of our proposed methods for each action, Table~\ref{tab:human3.6m-comparison-action} provides a comparison of P1 and P2 errors between our method and alternative methods on the Human3.6M dataset. Our proposed methods demonstrate superior performance in terms of P1 error across various actions such as Direction, Discuss, Eating, Greet, Purchase, Sitting, Sitting Down, Smoke, Wait, and Walk Dog, when compared to other existing methods. For the remaining actions, our methods deliver the second-best results.
    \begin{figure}[!t]
          \centering
          \includegraphics[width=0.5\textwidth]{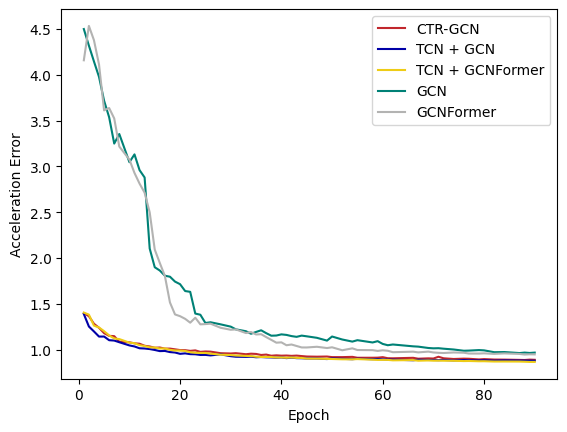}
          \caption{Comparison of acceleration error among different alternative modules designed for capturing local temporal dependencies.}
          \label{fig:acc-error-temp-gcnformer}
    \end{figure}
    \begin{figure*}[!t]
      \centering
      \includegraphics[width=17cm]{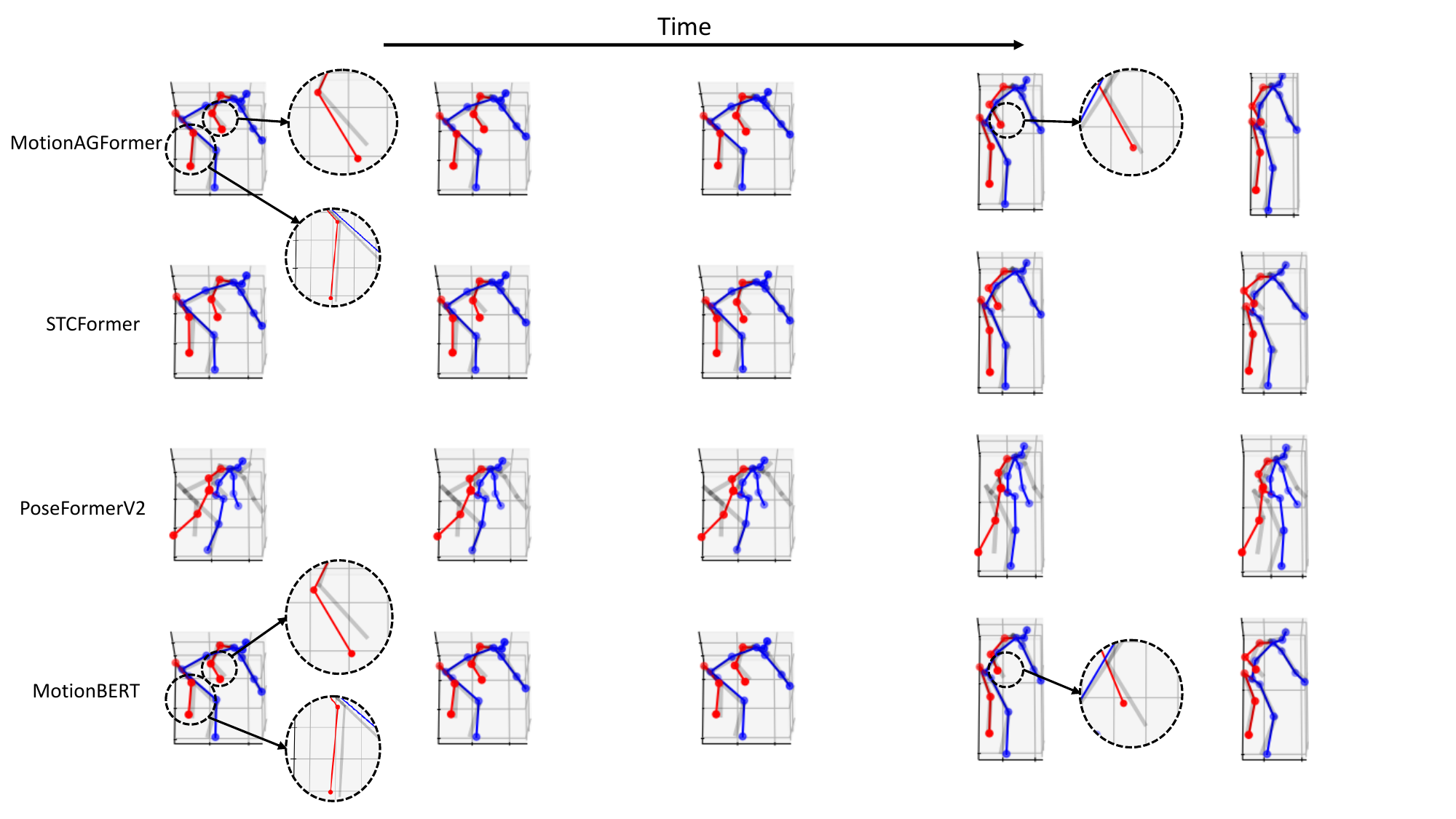}
      \caption{Qualitative comparisons of MotionAGFormer with STCformer~\cite{STCFormer}, PoseFormerV2~\cite{poseformerv2}, and MotionBERT~\cite{motionbert} on a random sequence from Human3.6M dataset.}
      \label{fig:sequence-comparison-1}
    \end{figure*}
    \begin{figure*}[!t]
      \centering
      \includegraphics[width=17cm]{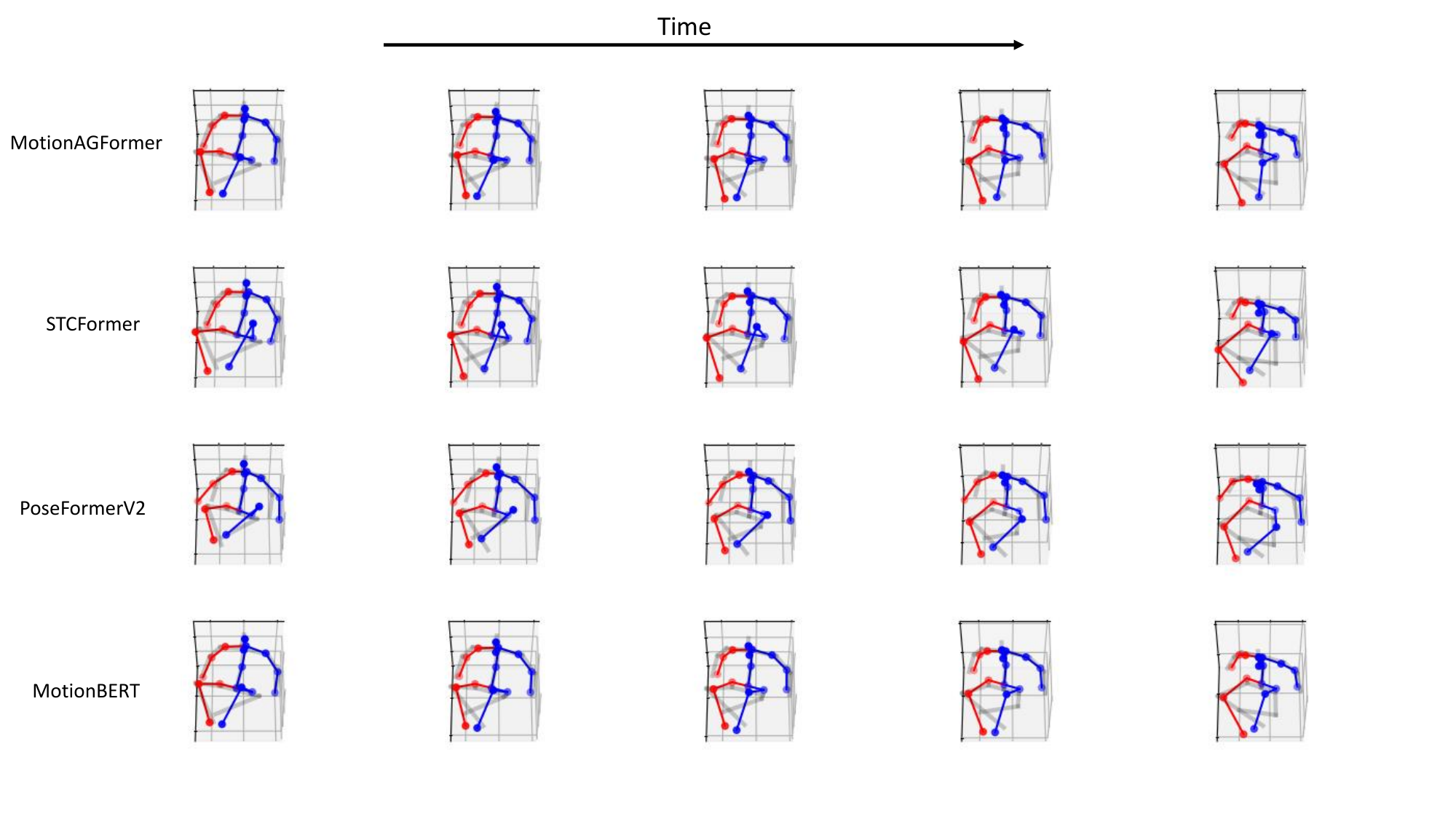}
      \caption{Failure case of the models on a random sequence from Human3.6M dataset.}
      \label{fig:sequence-comparison-2}
    \end{figure*}
    \subsection{Per-joint Error Comparison}
    Figure~\ref{fig:per-joint-error} shows per-joint comparisons of MotionAGFormer-B, MotionBERT, STCFormer, and StridedFormer on Human3.6M benchmark dataset. STCFormer demonstrates a slight performance advantage on hips, whereas for shoulders, MotionAGFormer exhibits a slight superiority over MotionBERT and significantly outperforms other models. MotionBERT generally outperforms MotionAGFormer on joints with a degree of freedom (DoF) of 2, including knees and elbows, with a notable gap between them and other models. However, for joints with a DoF of 3, such as ankles and wrists, MotionAGFormer consistently outperforms all other models, which is particularly important due to their greater contribution to overall error and their significance in applications like hand/object interaction and gait analysis.
    \section{Ablation on Temporal GCNFormer}
    In the main manuscript, we examined various alternatives for the temporal GCNFormer and reported their corresponding P1 errors. Nevertheless, we have yet to investigate the acceleration error associated with each of these alternatives. The acceleration error is defined as follows:
    \begin{equation} \label{acc-error}
    \begin{aligned}
      Acc(\mathbf{P}, \mathbf{\hat{P}}) &= \Sigma_{t=1}^{T-1} L(\mathbf{P}, t) - L(\mathbf{\hat{P}}, t), \\
      L(\mathbf{X}, t) &= \mathbf{X}_{t-1} - 2\mathbf{X}_{t} + \mathbf{X}_{t+1},
    \end{aligned}
    \end{equation}
    where $\mathbf{P}$ and $\mathbf{\hat{P}}$ denote the ground truth 3D pose and estimated 3D pose, respectively.

    Figure~\ref{fig:acc-error-temp-gcnformer} illustrates the acceleration error of different models when used to capture temporal local dependencies. Interestingly, despite CTR-GCN and TCN resulting in higher P1 error (as mentioned in the main manuscript), their acceleration error convergence during training is faster. In addition, it leads to an eventual acceleration error improvement of approximately 0.1\,mm.
    \section{Qualitative comparison for a sequence}
    Apart from making qualitative comparisons using a random frame from the Human3.6M dataset, we also analyzed consecutive frames within a single sequence. This allows us to observe the behavior of the compared models over a sequence, rather than qualitatively evaluating them solely based on a single frame. Figure~\ref{fig:sequence-comparison-1} compares MotionAGFormer-B with STCFormer, PoseFormerV2, and MotionBERT. Overall, MotionAGFormer and MotionBERT demonstrate superior alignment with the ground truth compared to the other two models. When comparing MotionAGFormer and MotionBERT, MotionAGFormer exhibits slightly better performance on joints with a DoF, such as wrists and ankles, as opposed to MotionBERT. There are also challenging sequences (see Figure~\ref{fig:sequence-comparison-2}) that all models fail to accurately estimate the joints.
    \section{In-the-wild Evaluation}
\begin{figure*}[!t]
          \centering
          \includegraphics[width=17cm]{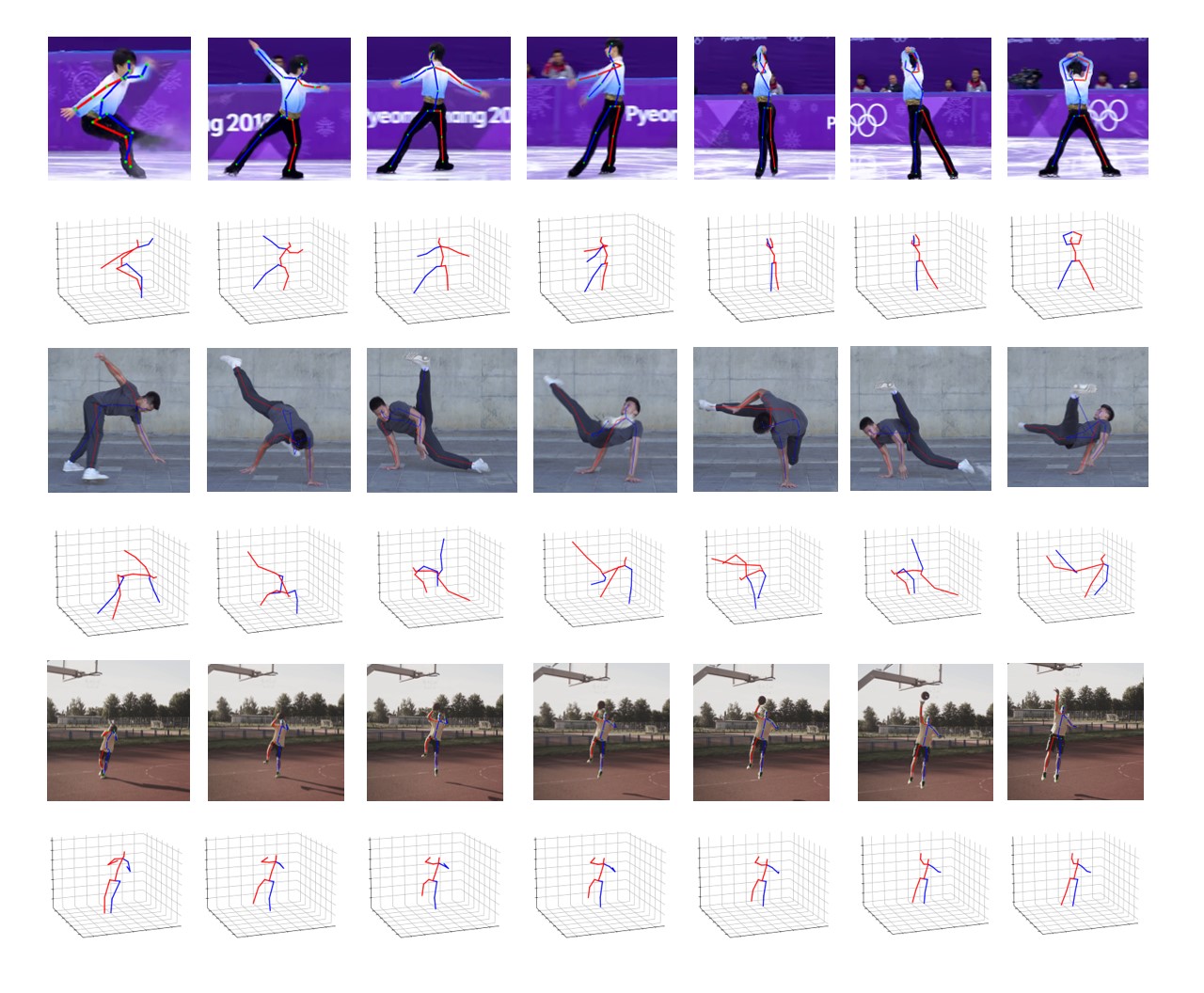}
          \caption{In-the-wild video evaluation of MotionAGFormer. Estimations are based on our base variant.}
          \label{fig:in-the-wild}
    \end{figure*}

    To verify the generalization of the proposed MotionAGFormer, we tested our base variant on some unseen in-the-wild videos. Our estimations show that even in scenarios where the inferred 2D poses exhibit some noise (as observed in the final frame of the second example shown in Figure~\ref{fig:in-the-wild}), the corresponding 3D pose estimations remain reasonable and resilient against sudden noisy movements in the 2D sequence. Moreover, when the subject is positioned at a distance from the camera, the 3D pose estimates exhibit a high degree of accuracy (as demonstrated in the last example of Figure~\ref{fig:in-the-wild}).
    
\end{document}